\documentclass[accepted]{uai2023} 

\usepackage[american]{babel}

\usepackage[space]{cite}
\usepackage{amsmath,amssymb,amsfonts}
\usepackage{algorithmic}
\usepackage{graphicx}
\usepackage{textcomp}
\usepackage{xcolor}
\usepackage{dirtytalk}
\usepackage{algorithm2e}
\usepackage{hyperref}
\usepackage{bm}
\usepackage{float}

\DeclareMathOperator*{\argmax}{arg\,max}

\newcommand{\Dcal}{\mathcal{D}}

\newtheorem{definition}{Definition}
\newtheorem{Theorem}{Theorem}

\newtheorem{Corollary}{Corollary}

\newtheorem{Hypothesis}{Hypothesis}

\usepackage{mathtools} 
\usepackage{booktabs} 
\usepackage{tikz} 

\title{
Approximately Bayes-Optimal Pseudo Label Selection}

\author[1]{\href{mailto:<rodemann@stat.uni-muenchen.de>?Subject=Your paper on Bayes Optimal Pseudo-Label Selection}{Julian Rodemann}{}}
\author[1,2,4]{Jann Goschenhofer}
\author[1,2,3]{Emilio Dorigatti}
\author[1,2]{Thomas Nagler}
\author[1]{Thomas Augustin}

\affil[1]{%
    Department of Statistics\\
    Ludwig-Maximilians-Universität (LMU)\\
    Munich, Germany
}
 \affil[2]{%
     Munich Center for Machine Learning (MCML)\\
     Munich, Germany
 }
  \affil[3]{%
     Institute of Computational Biology\\
     Helmholtz Zentrum\\
     Neuherberg, Germany
 }
   \affil[4]{%
     Fraunhofer Institute of Integrated Circuits (IIS)\\
     Erlangen, Germany
 }
  
\begin{document}
\maketitle

\begin{abstract}
  Semi-supervised learning by self-training heavily relies on pseudo-label selection (PLS). 
  The selection often depends on the initial model fit on labeled data. Early overfitting might thus be propagated to the final model by selecting instances with overconfident but erroneous predictions, often referred to as confirmation bias. 
  This paper introduces BPLS, a Bayesian framework for PLS that aims to mitigate this issue. 
  At its core lies a criterion
  for selecting instances to label: an analytical approximation of the posterior predictive of pseudo-samples. 
  We derive this selection criterion by proving Bayes optimality of the posterior predictive of pseudo-samples.
  We further overcome computational hurdles by approximating the criterion analytically. 
  Its relation to the marginal likelihood allows us to come up with an approximation based on Laplace's method and the Gaussian integral.
We empirically assess BPLS for parametric generalized linear and non-parametric generalized additive models \mbox{on} simulated and real-world data. When faced with high-dimensional data prone to overfitting, BPLS outperforms traditional PLS methods.\footnote{Code available at: \url{https://anonymous.4open.science/r/Bayesian-pls}}
\end{abstract}


\section{INTRODUCTION}
\label{sec:intro}

Labeled data are scarce in many learning settings. This can be due to a variety of reasons such as restrictions on time, knowledge, or financial resources. Unlabeled data, however, are often much more accessible. This has given rise to the paradigm of semi-supervised learning (SSL), where information from unlabeled data is integrated into model training to improve predictions in a supervised learning framework. 
Within SSL, an intuitive and widely used approach is referred to as self-training or pseudo-labeling \cite{shi2018transductive, lee2013pseudo, mcclosky2006effective}. The idea is to fit an initial model to labeled data and iteratively assign pseudo-labels to unlabeled data according to the model's predictions. 
The latter requires a criterion (sometimes called confidence measure) for pseudo-label selection (PLS), that is, the selection of instances to be pseudo-labeled and added to the training data. 

By design, self-training strongly relies on the initial model fit and the way instances are selected to be pseudo-labeled. Everything hinges upon the interplay \mbox{between} the selection criterion and the initial model's generalization performance. 
If the initial model generalizes poorly, initial misconceptions can propagate throughout the process, only making things worse. High-dimensional data prone to overfitting are particularly sensitive to such confirmation bias \cite{arazo2020pseudo}. Usually, self-training's sweet spot lies somewhere else: When the labeled data allow the model to learn sufficiently well while still leaving some room for improvement.  
Generally, the poorer the initial generalization, the harder it is to select sensible pseudo-labels to improve generalization, i.e., the more crucial the role of the selection criterion. 
Note that SSL is applied to data with high shares (typically over $80\%$ \cite{Sohn2020, arazo2020pseudo}) of unlabeled data, where initial overfitting is likely for high-dimensional models, while final overfitting is not.

\subsection{Motivation}

Accordingly, we strive for a selection criterion that is robust with respect to the initial model fit, i.e., its learned parameters. At the same time, it should still exploit the information in the labeled data. Such a measure calls for disentangling the uncertainty contributions of the data and the model's parameters. This is in line with recent work in uncertainty quantification (UQ) that suggests decomposing epistemic uncertainty into approximation uncertainty driven by (a lack of) data and modeling uncertainty driven by (primarily parametric) assumptions \cite{hullermeier2021aleatoric}. Bayesian inference offers a sound and consistent framework for this distinction. Its rationale of technically modeling not only data but also parameters as random variables has proven to offer much insight into UQ for machine learning \cite{hullermeier2021aleatoric} and deep learning \cite{abdar2021review, malinin2018predictive}. 

We exploit the Bayesian framework for pinpointing uncertainty with regard to data and parameters in PLS. Our approach of Bayesian pseudo-label selection (BPLS) enables us to choose pseudo-labels that are likely given the observed labeled data but not necessarily likely given the estimated parameters of the fitted model. 
What is more, BPLS allows to include prior information not only for predicting but also for selecting pseudo-labels. Notably, BPLS is flexible enough to be applied to any kind of predictive model whose likelihood and Fisher-information are accessible, including non-Bayesian models. BPLS entails a Bayes optimal selection criterion, the \textit{pseudo posterior predictive} (PPP). Its intuition is straightforward yet effective: By averaging over all parameter values, PPP is more robust towards the initial fit compared to the
predictive distribution based on a single optimal parameter vector. Our approximate version of the PPP is simple and computationally cheap to evaluate: $ \ell (\hat \theta) - \frac{1}{2} \log \lvert \mathcal{I(\hat \theta)} \rvert $ with $\ell(\hat \theta)$ being the likelihood and $\mathcal{I(\hat \theta)}$ the Fisher-information matrix at the fitted parameter vector $\hat \theta$. As an approximation of the joint PPP, it does not require an \textit{i.i.d.} assumption, rendering it applicable to a wide range of applied learning setups.

\subsection{Main Contributions}

\textbf{(1)} We derive PPP by formalizing PLS as a decision problem and show\footnote{Proofs of all theorems in this paper can be found in the supplementary material.} that PPP corresponds to the Bayes criterion, rendering selecting instances with regard to it Bayes optimal, see sections~\ref{sec:case-for-margl} and~\ref{sec:bayes-opt}.

\textbf{(2)} Since our selection criterion includes a possibly intractable integral, we provide analytical approximations, exploiting Laplace's method and the Gaussian integral, both for uninformative and informative priors. Using varying levels of accuracy, we balance the trade-off between computational feasibility and precision, see Section~\ref{sec:approx}. 

\textbf{(3)} We provide empirical evidence\footnote{
Implementations of the proposed methods as well as reproducible scripts for the experiments are provided in the anonymous repository named \textbf{Bayesian-pls} (\say{\textit{Bayesian, please!}}), see abstract. 
} for BPLS' superiority over traditionally predominant PLS methods in case of semi-supervised generalized additive models (GAMs) and generalized linear models (GLMs) faced with high-dimensional data prone to overfitting, see Section~\ref{sec:results}.





\section{BAYESIAN PLS}
\label{sec:bpls}

Most semi-supervised methods deal with classification or clustering tasks \cite{van2020survey, chapelle2006semi}. Loosely leaning on \cite{triguero2015self}, we formalize SSL as follows. Consider labeled data \begin{equation}
    \mathcal{D}=\left\{\left(x_{i}, y_{i}\right)\right\}_{i=1}^{n} \in \left(\mathcal{X} \times \mathcal{Y}\right)^{n}
\end{equation} and unlabeled data 
\begin{equation}
    \mathcal{U}=\left\{\left(x_{i}, \mathcal{Y}\right)\right\}_{i=n+1}^{m} \in \left(\mathcal{X} \times 2^\mathcal{Y}\right)^{m-n}
\end{equation} from the same data generation process, where $\mathcal{X}$ is the feature space and $\mathcal{Y}$ is the categorical target space. The aim of SSL is to learn a predictive classification function $f$ such that $f( x) = \hat y \in \mathcal{Y}$ utilizing both $\mathcal{D}$ and $\mathcal{U}$.  
As is customary in self-training, we start by fitting a model with unknown parameter vector $\theta \in \Theta$, $\Theta$ compact with $\dim(\Theta) = q$, on labeled data $
    \mathcal{D}=\left\{\left(x_{i}, y_{i}\right)\right\}_{i=1}^{n} $.
Our goal is -- as usual -- to learn the conditional distribution of $p( y \mid x)$ through $\theta$ from observing features ${x} = (x_1, \dots, x_n) \in \mathcal{X}^n$, and responses ${y} = (y_1, \dots, y_n) \in \mathcal{Y}^n$ in $\mathcal{D}$. 
Adopting the Bayesian perspective, we can state a prior function over $\theta$ as $\pi(\theta)$. The prior can represent information on $\theta$ but may also be uninformative. 

Within existing frameworks for self-training (see Section~\ref{sec:background}) in SSL, one could deploy such a Bayesian setting for \textit{predicting} unknown labels of $\mathcal{U}=\left\{\left(x_{i}, \mathcal{Y}\right)\right\}_{i=1}^{m}$ as well as for the final predictions on unseen test data. However, we aim at a Bayesian framework for \textit{selecting} pseudo-labels. This is beneficial for two reasons. First and foremost, considering the Bayesian posterior predictive distribution in PLS will turn out to be more robust towards the initial fit on $\mathcal{D}$ than classical selection criteria. Second, the Bayesian engine brings along the usual benefit of allowing to explicitly account for prior knowledge when selecting instances to be labeled. Notably, our framework of Bayesian pseudo-label \textit{selection} is unrelated to how pseudo-labels are \textit{predicted}. 

\subsection{The Case for the Posterior Predictive in PLS}
\label{sec:case-for-margl}

For any model with parameters $\theta \in \Theta$, the likelihood function for observed features ${x}$ and labels ${y}$ is commonly defined as 
$
    \mathcal {L}_{ y \mid  x}(\theta) = f_{\theta }( y\mid  x),
$ where $f_{\theta }(\cdot)$ is from a parameterized family of probability density functions. 
In the Bayesian universe, parameters $\theta$ are more than just functional arguments \cite{murphy2012machine}. They are random quantities themselves, allowing us to condition on them: $\mathcal{L}_{ y \mid  x}(\theta) = p({y}\mid {x}, \theta )$. Recall that we have specified a prior $\pi(\theta)$ on the parameters beforehand. After observing data, it can be updated to a posterior following Bayes' Theorem 
    $p(\theta \mid {y}, {x}) = p({y} \mid {x}, \theta ) \, \pi(\theta) / p({y} \mid {x}),$
where the denominator is the marginal likelihood 
\begin{equation}
p({y} \mid {x}) = \int_{\Theta} p({y} \mid {x}, \theta ) \, \pi(\theta )\,d\theta,
\end{equation} 
or, more colloquially, \say{Bayesian evidence} \cite{lotfi2022bayesian, barber2012bayesian}. For previously unseen data $(\Tilde{y}, \Tilde{x})$, the posterior predictive distribution is defined as

\begin{equation}
\label{eq:pp}
    p(\Tilde{y} \mid \Tilde{x}, {y}, {x})=\int_{\Theta} p(\Tilde{y} \mid \Tilde{x}, \theta) \, p(\theta \mid {y}, {x})\,d\theta.
\end{equation}

The posterior predictive closely resembles the marginal likelihood in case we include $(\Tilde{y}, \Tilde{x})$ in the data -- a fact that we will exploit for our approximations in Section~\ref{sec:approx}. Both marginalize the likelihood over $\theta$. The difference is the weight: The marginal likelihood integrates out $\theta$ with regard to the prior, while the posterior predictive integrates out $\theta$ with regard to the posterior. Accordingly, both can be considered PLS criteria that are robust towards the initial fit: They average over all possible $\theta$-values instead of relying on one estimated $\hat \theta$ from the trained model. 
\footnote{The probabilistic interpretation of the marginal likelihood -- in the words of \cite{lotfi2022bayesian} -- is: \say{The probability that we would generate a dataset with a model if we randomly sample from a prior over its parameters}. The posterior predictive, analogously, is the probability that we would generate data with a model if we randomly sample from a posterior over its parameters.}
Computational issues aside, the posterior predictive of pseudo-labeled data thus encapsulates a perfectly natural selection criterion for self-training: It selects pseudo-labels that are most likely conditioned on the true observed $\mathcal{D}$, the assumed model and all plausible parameters from the prior or posterior, respectively.



Both the data and the estimated parameters (as functions of the data) will change throughout the process of self-training. We argue that conditioning the choice of unlabeled instances solely on the estimated parameters in early iterations over-emphasizes the influence of the initial model. This optimistic reliance can be harmful in case of small $n$ and high $q$, where overfitting is likely. Selecting instances by the posterior predictive mitigates this.

\subsection{Bayes Optimality of Pseudo Posterior Predictive}
\label{sec:bayes-opt}

In the following, we show that selecting pseudo-labels with regard to their posterior predictive is Bayes optimal. We further show the same holds for selection with regard to the marginal likelihood in case of a non-updated prior. To this end, we formalize the selection of data points to be pseudo-labeled as a canonical decision problem, where an action corresponds to the selection of an instance from the set of unlabeled data $\mathcal{U}$.   

\begin{definition}[PLS as Decision Problem]
\label{def:dec-probl}    
Consider the decision-theoretic triple $(\mathcal{U}, \Theta, u(\cdot))$ with an action space of unlabeled data\footnote{We assume absence of tied observations for simplicity such that we can understand $\mathcal{U}$ as set.} to be selected, i.e., instances $(x_i, \mathcal{Y})$ as actions, a space of unknown states of nature (parameters) $\Theta$ and a utility function $u : \mathcal{U} \times \Theta \to \mathbb{R}$. 
\end{definition}

Loosely inspired by \cite{cattaneo2007statistical}, we now define the utility of a selected data point $(x_i, \mathcal{Y})$ as the plausibility of being generated jointly with $\mathcal{D}$ by a model with parameters $\theta \in \Theta$ if we include it with pseudo-label $\hat y_i \in \mathcal{Y}$ (obtained through any predictive model) in $\mathcal{D} \cup (x_i, \hat{y}_i)$. This is incorporated by the likelihood of $\mathcal{D} \cup (x_i, \hat{y}_i)$, which shall be called \textit{pseudo-label likelihood} and written as $p(\mathcal{D} \cup (x_i, \hat{y}_i) \mid \theta)$. We thus condition the selection problem on a model class as well as on already predicted pseudo-labels. The former conditioning is not required (see the extension in Section~\ref{sec:ext}) for the well-definedness of the pseudo-likelihood while the latter is. 

\begin{definition}[Pseudo-Label Likelihood as Utility]
\label{def:pseud-lik}
Let~$(x_i, \mathcal{Y})$ be any decision (selection) from $\mathcal{U}$. We assign utility to each $(x_i, \mathcal{Y})$ given $\mathcal{D}$ and pseudo-labels $\hat{y} \in \mathcal{Y}$ by the following measurable utility function
\begin{align*}
  u \colon \mathcal{U} \times \Theta &\to \mathbb{R}\\
  ((x_i, \mathcal{Y}), \theta) &\mapsto u((x_i, \mathcal{Y}), \theta) = p(\mathcal{D} \cup (x_i, \hat{y}_i)\mid \theta),
  \end{align*}
  which is said to be the pseudo-label likelihood.
\end{definition}

This utility function is a natural probabilistic choice to assign utilities to selected pseudo-labels given the predicted pseudo-labels. With a prior $\pi(\theta)$, we get the following result.

\begin{Theorem}
\label{th:bayes-opt}
In the decision problem $(\mathcal{U}, \Theta, u(\cdot))$ (Definition~\ref{def:dec-probl}) with the pseudo-label likelihood as utility function (Definition~\ref{def:pseud-lik}) and a prior $\pi(\theta)$ on $\Theta$, the standard Bayes criterion 
\begin{align*}
    \Phi(\cdot,\pi) \colon \mathcal{U} &\to \mathbb{R}\\
    a &\mapsto \Phi(a, \pi) = \mathbb{E}_\pi(u(a,\theta)) 
\end{align*}

corresponds to the pseudo marginal likelihood $p(\mathcal{D}~\cup~(x_i, \hat{y}_i))$.
    
\end{Theorem}


\begin{Corollary}
   For any prior $\pi(\theta)$ on $\Theta$, the action $a_{m}^* = \argmax_i p(\mathcal{D} \cup (x_i, \hat{y}_i))$ is Bayes optimal.
\end{Corollary}

Taking the observed labeled data $\mathcal{D}$ into account by updating the prior $\pi(\theta)$ to a posterior $p(\theta \mid \mathcal{D})$, we end up with an analogous result for the \textit{pseudo posterior predictive}. The Theorem requires only the Proposition by \cite[section 4.4.1]{berger1985statistical} stating that posterior loss equals prior risk. That is, conditional Bayes optimality equals unconditional Bayes optimality.  

\begin{Theorem}
\label{th:ppp}
In the decision problem $(\mathcal{U}, \Theta, u(\cdot))$ and the pseudo-label likelihood as utility function as in Theorem~\ref{th:bayes-opt} but with the prior updated by the posterior $\pi(\theta) = p(\theta \mid \mathcal{D})$ on $\Theta$, the standard Bayes criterion 
$\Phi(\cdot, \pi) \colon \mathcal{U} \to \mathbb{R}; \, a \mapsto \Phi(a, \pi) = \mathbb{E}_\pi(u(a,\theta)) $
corresponds to the \textit{pseudo posterior predictive} $p(\mathcal{D} \cup (x_i, \hat{y}_i)\mid \mathcal{D})$.
\end{Theorem}


\begin{Corollary}
   Action $a_{p}^* = \argmax_i p(\mathcal{D} \cup (x_i, \hat{y}_i)\mid \mathcal{D})$ is Bayes optimal for any updated prior $\pi(\theta) = p(\theta \mid \mathcal{D})$.
\end{Corollary}


Further note that directly maximizing the likelihood with regard to $a$ corresponds to the optimistic max-max-criterion, see Theorem~\ref{th:max-max}.

\begin{Theorem}
\label{th:max-max}
In the decision problem $(\mathcal{U}, \Theta, u(\cdot))$ with the pseudo-label likelihood as utility function as in Theorem~\ref{th:bayes-opt}, the max-max criterion 
\begin{align*}
    \Phi(\cdot) \colon \mathcal{U} &\to \mathbb{R};\\
    a &\mapsto \Phi(a) = \max_\theta (u(a,\theta)) 
\end{align*}

corresponds to the (full) likelihood.
\end{Theorem}

The max-max-criterion advocates deciding for an action (here: selection of pseudo-labeled data) with the highest utility (here: likelihood) according to the most favorable state of nature $\theta$, e.g. see \cite{rapoport1998decision}. It can hardly be seen as a rational criterion, as it reflects \say{wishful thinking} \cite[page 57]{rapoport1998decision}. We thus abstain from it in what follows. Our roughest approximation of the PPP in Section~\ref{sec:approx}, however, will correspond to this case as well as the more general concept of optimistic superset learning (OSL) \cite{hullermeier2014learning, rodemann2022levelwise}.



\section{Approximate Bayes Optimal PLS}
\label{sec:approx}

Since the \textit{pseudo posterior predictive} (PPP) $p(\mathcal{D} \cup (x_i, \hat{y}_i)\mid \mathcal{D}) = p(\hat{y} \mid x,  y,  x)$ (Theorem~\ref{th:ppp}) is computationally costly to evaluate via Markov Chain Monte Carlo (MCMC), we aim at approximating it analytically. In light of the general computational complexity of BPLS involving model refitting, see Section~\ref{sec:algo}, this appears particularly crucial.
We will approximate the joint PPP directly.\footnote{  
For \textit{i.i.d.} data we could focus on the single PPP contributions $p(y_i \mid x_{i}, \mathcal{D})$ instead of the joint. Still, we would have to deal with a possibly intractable integral and end up with similar computational hustle. We thus opt for approximating the joint directly. Moreover, considering the joint quantities instead of the distributions implies no loss of generality, with possible extensions for dependent data in mind.} Our method hence does not need an \textit{i.i.d.} assumption, which makes it very versatile. 

Due to the aforementioned similarity of the PPP and the marginal likelihood, we are in the fortunate position of borrowing from some classical marginal likelihood approximations, see \cite{llorente2020marginal}. Especially popular are approximations based on Laplace's method as in \cite{schwarz1978estimating}. 
Our main motivation, however, is to obtain a Gaussian integral \cite{gauss1877theoria}, which we can then compute explicitly.

\subsection{Approximation of the PPP}

We will start by transferring Laplace's method to the PPP. 
Recall that the predictive posterior of a pseudo-sample $(x_i, \hat y_i)$ (the PPP) given data $\Dcal$ is defined as 
\begin{align*}
    p(\Dcal \cup (x_i, \hat y_i) | \Dcal)  &=  \int_\Theta  p(\Dcal \cup (x_i, \hat y_i) \mid \theta) p(\theta \mid \Dcal) d\theta,
\end{align*}
where Bayes' theorem gives
\begin{align*}
  p(\theta \mid \Dcal) =  p(\Dcal \mid \theta) \pi(\theta) / p(\Dcal).
\end{align*}
Denoting $\ell_{\Dcal}(\theta) = \log p(\Dcal \mid \theta)$ and $\tilde \ell(\theta) = \ell_{\Dcal\cup (x_i, \hat y_i)}(\theta)  + \ell_{\Dcal}(\theta)$, we can write the integrand as
\begin{align*} 
  p(\Dcal \cup (x_i, \hat y_i) \mid \theta) p(\theta \mid \Dcal) = \exp[\tilde \ell(\theta) \bigr) ] \pi(\theta) / p(\Dcal).
\end{align*}
Let $\mathcal I(\theta) = -\tilde \ell''(\theta)/n$ denote the observed Fisher information matrix. Further denote by  $\tilde \theta = \arg\max_{\theta} \tilde \ell(\theta)$ the maximizer of $\tilde \ell(\theta)$. It holds $\tilde \ell'(\theta) = 0$ by definition of $\tilde \theta$. A Taylor expansion around $\tilde \theta$ thus gives
\begin{align*} 
    \tilde \ell(\theta) \approx \tilde \ell(\tilde \theta)   - \frac{n}{2} (\theta - \tilde  \theta)' \mathcal{I}(\tilde  \theta) (\theta - \tilde \theta).
\end{align*}
The integrand decays exponentially in $n\|\theta - \tilde \theta\|$, so we can approximate it locally around $\tilde \theta$ by also taking $\pi(\theta)~\approx~ \pi(\tilde \theta)$ inside the integral with an analogous Taylor series. We refer to \cite[Section 3.7]{miller2006applied} and \cite[Theorem 2]{lapinski2019} for a rigorous treatment of the remainder terms and regularity conditions. 

We can eventually approximate $p(\Dcal \cup (x_i, \hat y_i) | \Dcal)$ by
\begin{align*} 
     \frac{\exp[\tilde \ell(\tilde \theta)] \pi(\tilde \theta)}{p(\mathcal{D})}  \int_{\Theta} \exp\biggl[- \frac{n}{2} (\theta - \tilde  \theta)'  \mathcal{I}(\tilde  \theta) (\theta - \tilde \theta)\biggr] d \theta,
\end{align*}
 The integral on the right is a Gaussian integral. Defining $\Sigma = [n \mathcal{I}(\tilde  \theta)]^{-1}$ and $\phi_\Sigma$ as the density of the $\mathcal N(0, \Sigma)$ distribution, it equals
\begin{align*}
   (2\pi)^{q/2} |\Sigma|^{1/2} \int_{\Theta} \phi_\Sigma(\theta) d \theta 
    =\biggl(\frac{2\pi}{n}\biggr)^{q/2} |\mathcal I(\tilde \theta)|^{-1/2}.
\end{align*}
Altogether, we have shown that 
\begin{align} \label{eq:laplace}
     p(\Dcal \cup (x_i, \hat y_i) | \Dcal) 
    &\approx\biggl(\frac{2\pi}{n}\biggr)^{q / 2} \frac{\exp[\tilde \ell(\tilde \theta) ]  \pi(\tilde \theta)}{ | \mathcal I(\tilde \theta)  |^{1/2} p(\Dcal) }.
\end{align}

\subsection{Approximate selection criteria}

To find the pseudo-sample $(x_i, \hat y_i)$ maximizing the PPP, we can equivalently maximize its logarithm, i.e. maximize

\begin{equation*}    
\begin{split}
   \frac{q}{2} \, \log\biggl(\frac{2\pi}{n}\biggr) + \tilde \ell(\tilde \theta) + \log \pi(\tilde \theta) - \frac{1}{2} \log|\mathcal{I}(\tilde \theta)| - \log p(\mathcal{D}).   
\end{split} 
\end{equation*}

Dropping all terms that do not depend on $(x_i, \hat y_i)$ leads to the selection criterion 
\begin{align} \label{eq:psl-informative-0}
  \tilde \ell(\tilde \theta) - \frac 1 2 \log |\mathcal I(\tilde \theta)| + \log \pi(\tilde \theta).
\end{align}

The term
$$\tilde \ell(\theta) = \ell_{\Dcal\cup (x_i, \hat y_i)}(\theta)  + \ell_{\Dcal}(\theta),$$
quantifies how well the pseudo-sample $(x_i, \hat y_i)$ conforms with the data set $\Dcal$ given a parameter $\theta$, e.g. the optimal (argmax) parameter $\tilde \theta$ in Equation~\eqref{eq:psl-informative-0}. 
It is curious that samples in $\Dcal$ contribute twice to $\tilde \ell$, but $(x_i, \hat y_i)$ only once. However, this is irrelevant when comparing two pseudo-samples $(x_i, \hat y_i)$ and $(x_j, \hat y_j)$. To see this, we expand $\ell_{\Dcal}$ around its maximizer $\hat \theta$, so that $\ell_{\Dcal}(\tilde \theta) =  \ell_{\Dcal}(\hat \theta) + O(\|\hat \theta - \tilde \theta\|^2)$. Since  $\Dcal \cup (x_i, \hat y_i)$ and $\Dcal$ differ in only one sample, the difference $\hat \theta - \tilde \theta$ is of order $O(n^{-1})$. Thus,
$$ \tilde \ell(\theta) = \ell_{\Dcal\cup (x_i, \hat y_i)}(\theta) + \ell_{\Dcal}(\hat \theta) + O(n^{-2}).$$
The remainder is negligible compared to the other terms in \eqref{eq:psl-informative-0} and $\ell_{\Dcal}(\hat \theta)$ does not depend on the pseudo-sample $(x_i, \hat y_i)$. This suggests the simplified \emph{informative BPLS criterion}
\begin{align} \label{eq:psl-informative}
   \operatorname{iBPLS} = \ell_{\Dcal \cup (x_i, \hat y_i)}(\tilde \theta) - \frac 1 2 \log | \mathcal I(\tilde \theta)| + \log \pi(\tilde \theta).
\end{align}
Equivalence of \eqref{eq:psl-informative-0} and \eqref{eq:psl-informative} is verified numerically for small $n$ by experiments on real-world and simulated data in Supplement~F.

The ability to incorporate prior information into the selection is generally a strength of our criterion. By default, however, we cannot assume that such information is available. We can instead choose an uninformative prior where $\pi(\theta)$ is constant with respect to $\theta$. Recall that we assume $\Theta$ to be compact, which allows us to specify a uniform prior as uninformative prior. Then \eqref{eq:psl-informative} simplifies to the \emph{uninformative BPLS criterion} 
\begin{align} \label{eq:psl-uninformative}
    \operatorname{uBPLS} = \ell_{\Dcal \cup (x_i, \hat y_i)}(\tilde \theta) - \frac 1 2 \log | \mathcal I(\tilde \theta)|.
\end{align}

Our novel PLS criteria provide great intuition.

\begin{itemize}
    \item The first term is the joint likelihood of the pseudo-sample $(x_i, \hat y_i)$ and $\Dcal$ under the optimal parameter $\tilde \theta$. It measures how well the pseudo-sample complies with the previous model and previously seen data $\Dcal$. It tells the value of this joint likelihood at its maximum. Loosely speaking, this maximum height of the likelihood can be seen as a very rough approximation of the area under it, i.e., the integral with uniform weights.\footnote{Technically, we also need that $\lambda(\Theta) = 1$ with $\lambda$ a Lebesgue-measure for this interpretation.} 
    \item The second term penalizes high curvature of the pseudo-label likelihood function $\ell_{\Dcal \cup (x_i, \hat y_i)}(\theta)$ around its peak, since the Fisher-information is its second derivative. Due to the negative sign, the criterion prefers pseudo-samples that lead to flatter maxima of the likelihood. In line with recent insights into sharp and flat minima of loss surfaces \cite{dinh2017sharp, li2018visualizing, andriushchenko22a}, such a penalty can be expected to improve generalization. The lower the curvature, the more probability mass (area under the likelihood) is expected on $\Theta \setminus B_{\epsilon}(\tilde \theta)$ with $B_{\epsilon} = \{\theta \in \Theta \mid \|\theta - \tilde \theta \| < \epsilon \}$ an $\epsilon$-ball for fixed $\epsilon > 0$ around $\tilde \theta$ in the uninformative case. Intuitively, this corrects the very rough approximation of the area under the likelihood by the likelihood's maximal height, see above.  
    \item The third term in the informative BPLS criterion adjusts the selection for our prior beliefs $\pi$ about $\theta$. Here, the effect of $(x_i, \hat y_i)$ is only implicit, because it affects the maximizer $\tilde \theta$. The more likely the updated parameter $\tilde \theta$ is under $\pi$, the higher the PPP.
\end{itemize}

In summary, our approximation of the PPP 
grows in the absolute value of the likelihood's peak, decreases in its curvature at this point, and increases in the prior likelihood of the updated parameter.

When $n \to \infty$, the criteria iBPLS and uBPLS are dominated by the likelihood, thus
$$
    \log  p(\Dcal \cup (x_i, \hat y_i) | \Dcal)   \overset{n \to \infty}{\propto}  \,  \ell_{\Dcal \cup (x_i, \hat y_i)}(\tilde \theta).
$$
This approximation is computationally cheaper to evaluate, as it does not involve the Fisher-information. However, this comes at the cost of poor accuracy in case of small $n$. Selection with regard to this rough approximation of the PPP corresponds to selection with regard to the likelihood.
As pointed out in Section~\ref{sec:bayes-opt}, this corresponds to the overly optimistic max-max-criterion.

\begin{figure*}[t!]
\centering
\includegraphics[width=\textwidth]{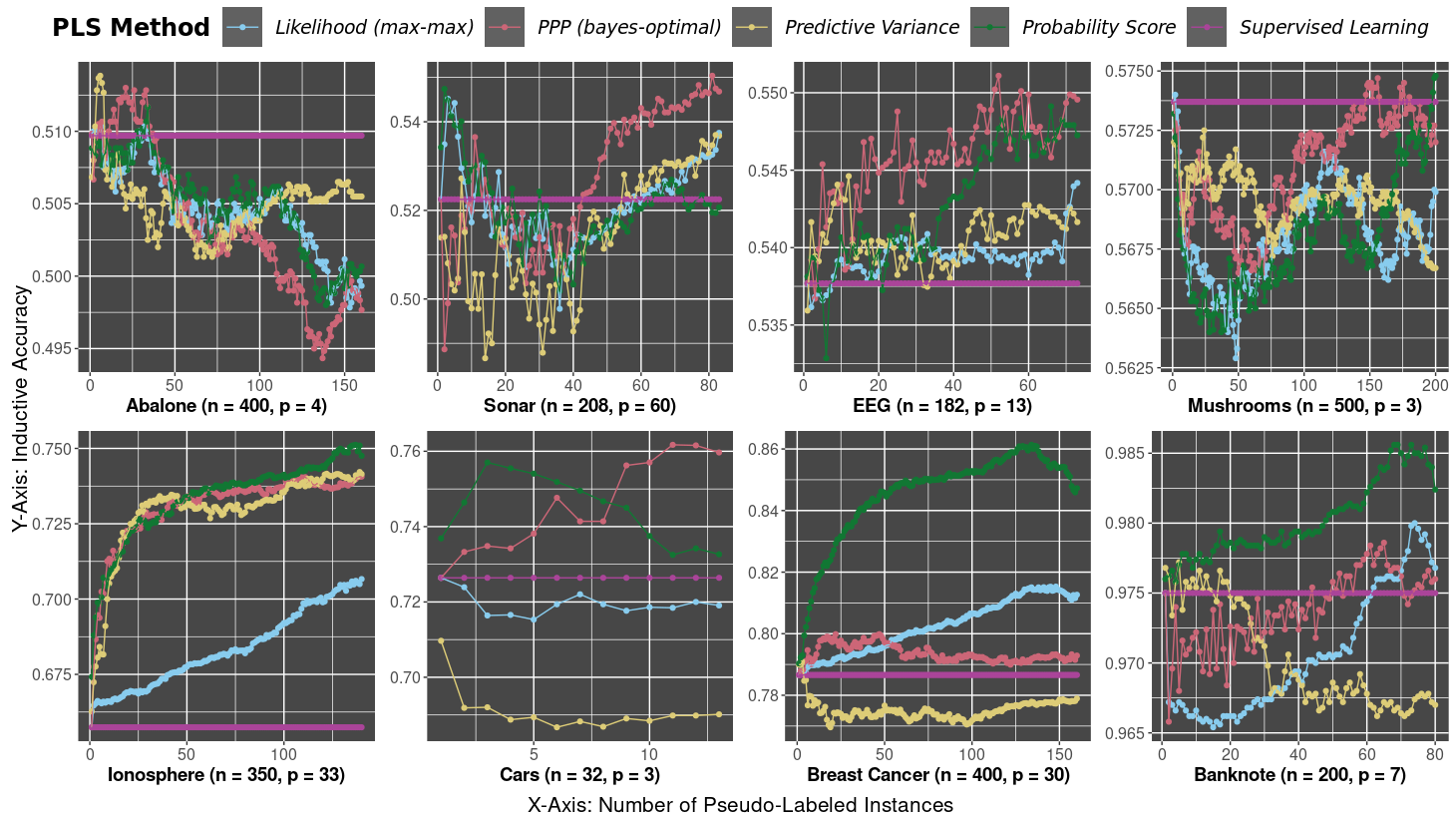}
\caption{Results from 8 classification tasks based on real-world data \cite{Dua:2019} in descending difficulty (measured by supervised test accuracy), where $p$ denotes the number of features here and the share of unlabeled data is 0.8. Accuracy averaged over 100 repetitions.}
\label{fig:all-results}
\end{figure*}


\section{EXPERIMENTS}
\label{sec:results}

\textbf{Algorithmic Procedure:}
\label{sec:algo}
For all predicted pseudo-labels, we refit the model on $\mathcal{D}~\cup~(x_i, \hat{y}_i)$ and evaluate its PPP by means of the derived approximations iBPLS and uBPLS to select one instance to be added to the training data. Detailed pseudo code for BPLS can be found in Supplement A. The computational complexity depends on the evaluation of the PPP. With $\lvert \mathcal{U} \rvert = m$ unlabeled data points and no stopping criterion, $m + (m - 1) + \dots + 1 = \frac{m^2 + m}{2}$ PPPs have to be evaluated (that is, approximated). Hence, BPLS' complexity depends on the model's complexity and the amount of unlabeled data.





\begin{Hypothesis}
\label{hypo:BPLS-good}
\textbf{(a)} PPP with uninformative prior outperforms traditional PLS on data prone to initial overfitting (i.e., with high ratio of features to data $\frac{q}{n}$ and poor initial generalization).   
\textbf{(b)} For low $\frac{q}{n}$ and high initial generalization, BPLS is outperformed by traditional PLS.
\end{Hypothesis}

\begin{Hypothesis}
\label{hypo:likelihood}
    \textbf{(a)} Among all PLS methods, the pseudo-label likelihood (max-max-action) reinforces the initial model fit the most and \textbf{(b)} hardly improves generalization. 
\end{Hypothesis}

\begin{Hypothesis}
\label{hypo:informative}
    PPP with informative prior outperforms traditional PLS methods universally.
\end{Hypothesis}

\textbf{Experimental Setup:} We formulate three hypotheses beforehand.
Hypothesis \ref{hypo:BPLS-good} corresponds to the main motivation behind BPLS; its second part is a logical consequence thereof: If we are sceptical towards the initial model in case it generalizes well, we expect to select pseudo-labels in a worse way than when trusting the initial model. Hypothesis \ref{hypo:likelihood} is based on the decision-theoretic insights regarding PLS by the likelihood, see section \ref{sec:bayes-opt}: It embodies an optimistic reliance on the initial model and is thus expected to pick data that fits best into that model. We further expect (Hypothesis \ref{hypo:informative}) BPLS to unambiguously outperform non-Bayesian selection methods in case the prior provides actual information about the data generating process -- the latter is simply not available for non-Bayesian PLS.
We benchmark semi-supervised (parametric) generalized linear models (GLMs) and (non-parametric) generalized additive models (GAMs) \cite{hastie1987generalized, hastie2017generalized} with PPP and pseudo-label likelihood against two common selection criteria (probability score and predictive variance) \cite{triguero2015self} as well as a supervised baseline. For the latter, we abstain from self-training and only use the labeled data for training. Experiments are run on simulated binomially distributed data as well as on eight data sets for binary classification from the UCI repository \cite{Dua:2019}. The binomially distributed data was simulated through a linear predictor consisting of normally distributed features. Details on the simulations as well as on the data sets can be found in Supplement~C and Supplement~H.
The share of unlabeled data was set to $0.8$ and $0.9$. PLS methods were compared w.r.t. to (\say{inductive}) accuracy of prediction on unseen test. All data sets were found to be fairly balanced except for the EEG data (minority share: $0.29$).

\begin{figure}
    \centering
    \includegraphics[width=\columnwidth]{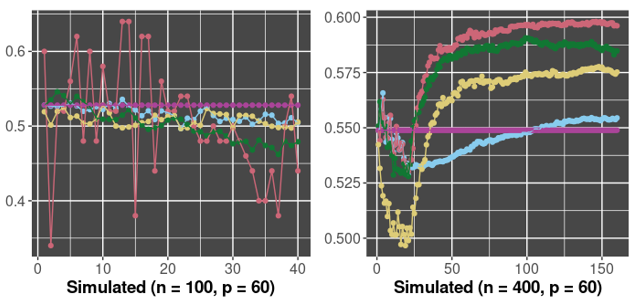}
    \caption{Results from simulated data. Accuracy averaged over 100 repetitions. Legend: see Figure \ref{fig:all-results}.}
    \label{fig:res-sim}
\end{figure}

\begin{figure*}[t!]
\centering
\includegraphics[width=\textwidth]{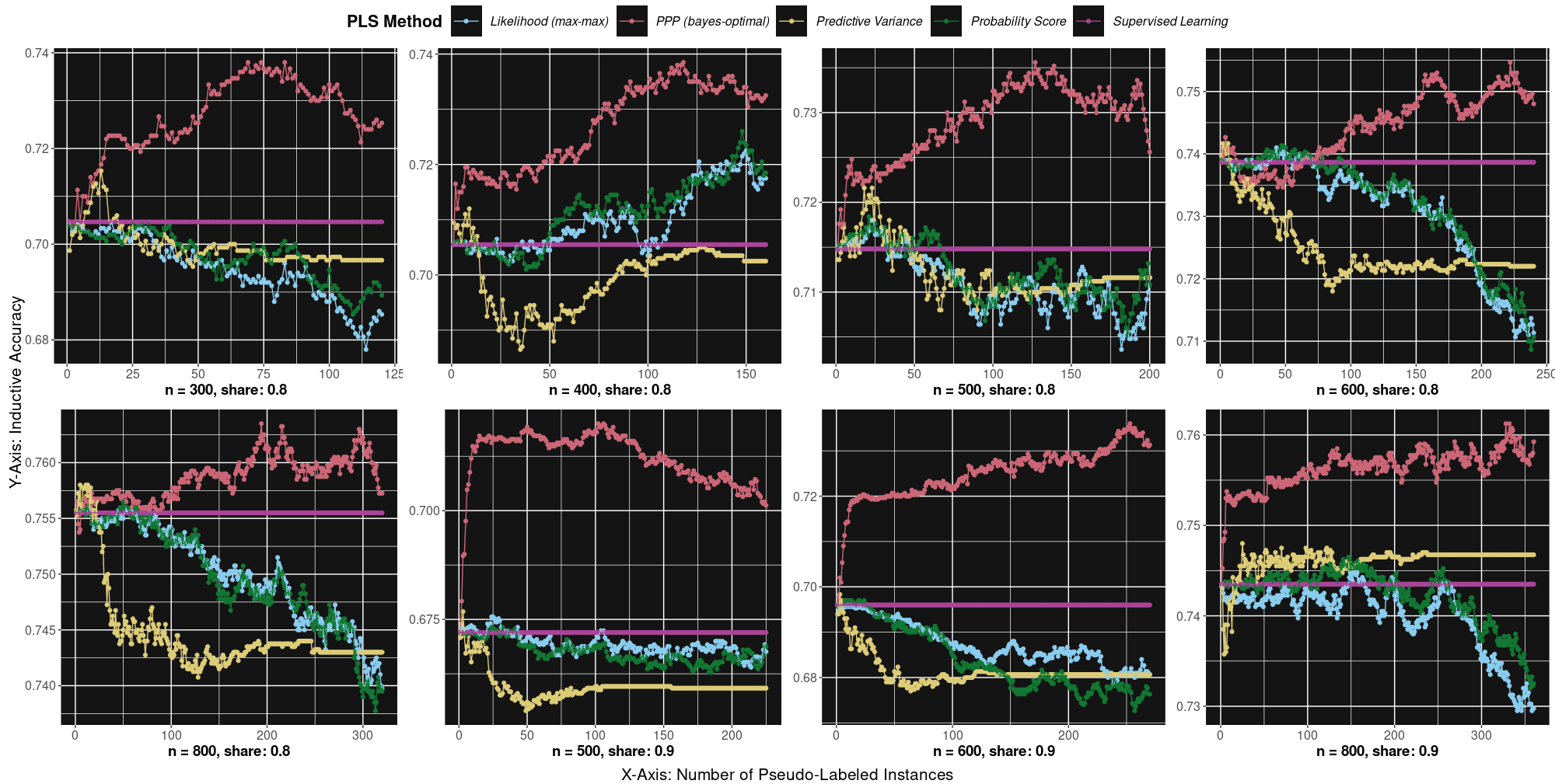}
\caption{Results of PPP with informative priors and non-parametric GAMs on simulated data with different shares of unlabeled data. Accuracy averaged over 100 repetitions.}
\label{fig:results-inf}
\end{figure*}

\textbf{Results:}
Figures~\ref{fig:all-results} and~\ref{fig:res-sim} as well as Table \ref{tab:table} summarize the results in the uninformative case (grey figures) for real-world and simulated data, respectively. \say{Oracle stopping} in Table~\ref{tab:table} refers to comparing PLS methods with regard to their overall best accuracy as opposed to \say{final} comparisons after the whole data set was labeled. Figure~\ref{fig:res-sim} sheds further light on results for simulated data, while Figure~\ref{fig:results-inf} displays results from benchmarking BPLS to classical PLS methods in the informative case (black figures). Detailed figures displaying results from all experiments can be found in the supplementary material.

\textbf{Interpretation:}
At first sight, comparing the accuracy gains in Figure~\ref{fig:all-results} on different data sets (in order of ascending baseline performance) clearly supports Hypothesis~\ref{hypo:BPLS-good}: For harder tasks like EEG or sonar with relatively high ratio of features to data $\frac{q}{n}$, Bayesian PPP outperforms traditional PLS, whilst being dominated by the probability score in case of easier tasks like banknote or breast cancer. For data sets with intermediate difficulty (mushrooms and ionosphere), PPP and other PLS methods compete head-to-head. The results on abalone data underpin a general fact in SSL (see section~\ref{sec:intro}): Successful self-training requires at least some baseline supervised performance.
Results on simulated data (Table~\ref{tab:table}) further support the role of $\frac{q}{n}$ in Hypothesis~\ref{hypo:BPLS-good}. Their visualization (Figure~\ref{fig:res-sim}) nicely illustrates the inner working of selection by PPP: By not trusting the initial model, PPP affects the model's test accuracy the most. While $n = 400$ leaves some room for improvement through mitigating the overfitting by pseudo-labeled data, PPP leads to a noisy performance in case of $n = 100$ close to $p$. Here, even the final model still overfits. These promising results should not hide an inconsistency: The fact that PPP is superior on the cars task but not on the ionosphere task contradicts Hypothesis~\ref{hypo:BPLS-good}, since cars is harder than ionosphere, while having almost identical $\frac{q}{n}$.
We find Hypothesis \ref{hypo:likelihood} to be partially supported by the results. While \ref{hypo:likelihood} (a) holds for both the majority of simulated (see supplementary material) and real-world data (likelihood generally the closest to supervised performance), \ref{hypo:likelihood} (b) is challenged by considerable generalization performance gain on ionosphere and breast cancer data. 

Figure \ref{fig:results-inf} clearly supports Hypothesis \ref{hypo:informative}: When using informative priors based on the true data-generating process, BPLS clearly outperforms traditional PLS methods. Results in Supplement D.3 further back this finding. This comes at no big surprise, since non-Bayesian PLS simply lack ways to incorporate such prior knowledge. From this perspective, the uninformative case (Hypothesis~\ref{hypo:BPLS-good}) corresponds to raising the bar and clearly is the theoretically more interesting benchmarking setup. However, many practical applications of SSL entail a myriad of pre-existing knowledge, e.g., radio spectrum identification \cite{cameloetal}. For practical purposes, thus, the informative situation might even be more relevant.

\begin{table}
\caption{Best performing PLS method (uninformative) on simulated data} 
\label{sample-table}
\begin{center}
\small
\begin{tabular}{c||c||ll}
\textbf{n} & \textbf{q} &\textbf{ORACLE STOPPING} & \textbf{FINAL} \\
\hline \hline
60 & 60 &  PPP         & PPP\\
100 & 60 & PPP & Supervised Learning \\
400 & 60 & PPP             & PPP \\
1000 & 60 & Probability Score & Probability Score \\
\end{tabular}
\end{center}
\label{tab:table}
\end{table}

\section{RELATED WORK}
\label{sec:background}

\textbf{Robust PLS:} Robustness of PLS is a widely discussed issue in the self-training literature. \cite{aminian2022information} propose information-theoretic PLS robust towards covariate shift. \cite{lienen2021credal} label instances in the form of sets of probability distributions (credal sets), weakening the reliance on a single distribution. \cite{vandewalle2013predictive} aim at robustness to modeling assumptions by allowing model selection through the deviance information criterion during semi-supervised learning. \cite{rizve2020defense} propose uncertainty-aware pseudo-label selection which proves to compete with state-of-the-art SSL based on consistency regulation. The idea is to select pseudo-labeled instances whose probability score and predictive uncertainty are above (tunable) thresholds. The latter is operationalized by the prediction's variance, and thus, unlike BPLS, fails to decompose approximation and modeling uncertainty, see Section~\ref{sec:intro}. Both predictive variance and probability score serve as benchmarks in Section~\ref{sec:results}.


\textbf{Bayesian Self-Training:} There is a broad body of research on deploying Bayesian \textit{predictions} in SSL and particularly in self-training \cite{gordon2020combining, ng2018bayesian, cai2011bassum, adams2009archipelago}. The same holds for explicit likelihood-based inference, such as weighted likelihood \cite{sokolovska2008asymptotics}, conditional likelihood \cite{grandvalet2004semi}, and joint mixture likelihood \cite{amini2002semi}. 
Most of them use Bayesian models for \textit{predicting} pseudo-labels. In contrast, we prove that the argmax of the PPP is the Bayes optimal \textit{selection} of pseudo-labels given \textbf{any} predictive model. 
Regarding Bayesian or likelihood-based \textit{selection} of pseudo-labels, there exists only little (Bayesian) or hardly any (likelihood-based) work. \cite{li2020pseudo} quantify the uncertainties of pseudo-labels by mixtures of predictive distributions of a neural net, applying MC dropout. This could be seen as an expensive MC-based approximation of the PPP. Very recently, \cite{patel2022seq} proposed PLS with regard to (a sampling-based approximation of) the entropy of the pseudo-labels' posterior predictive distribution. The entropy is considered a measure of total uncertainty (aleatoric and epistemic) and often considered as regularization for PLS, see \cite{saporta2020esl, liu2021cycle} for instance. Abstaining from the entropy -- as we do -- effectively means not considering the aleatoric uncertainty. While including aleatoric uncertainty (e.g. measurement noise) generally makes sense, we consider it of minor importance in the concrete problem of initial overfitting, where we aim at disentangling epistemic uncertainty with regard to \textit{data} and \textit{parameters}: We want to choose pseudo-labels that are likely given the \textit{observed labeled data} but not necessarily likely given the estimated \textit{parameters of the (over-)fitted model}.

\section{DISCUSSION}
\label{sec:discussion}

\textbf{Extensions:}
\label{sec:ext}
We briefly discuss two venues for future work. 
The first extension loosens the restriction on one particular model class by performing model selection and PLS simultaneously. The idea would be to select these instances that can be best explained by the simplest learner (i.e., the one with least parameters).  
Further recall that both the framework of BPLS and our approximation of the PPP do not require data to be \textit{i.i.d} distributed. Applying BPLS on dependent observations, such as in auto-correlated data like time series, is thus another promising line of further research.

\textbf{Limitations:}
BPLS' strength of being applicable to any learner can imply high computational costs in case of expensive-to-train models such as neural nets, because PPP approximations require refitting the model. Additionally, it might be difficult for practitioners to assess the risk of overfitting to the initial data set beforehand and opt for BPLS in response. Given the fact that BPLS is outperformed by traditional PLS in cases with no overfitting, this might be considered a drawback for practical application. However, Section~\ref{sec:results} demonstrated that $\frac{q}{n}$ and the baseline supervised performance (both easily accessible) provide sound proxies for initial overfitting scenarios that can induce a confirmation bias in PLS. These proxies can (alongside cross-validation) help practitioners to identify such scenarios.

\textbf{Conclusion:}
BPLS renders self-training more robust with respect to the initial model. This improves final performance if the latter overfits and harms it if not. Identifying overfitting scenarios is thus crucial for BPLS' usage. 
What is more, BPLS allows incorporating prior knowledge, with the help of which substantial performance gains can be achieved. Besides, our insights from formalizing PLS as a decision problem clear the way for promising future work exploiting rich literature on Bayesian decision theory. Ultimately, we conclude that a Bayesian view can add great value not only to predicting but also to selecting data for self-training.

\subsubsection*{Acknowledgements}

Thomas Augustin gratefully acknowledges support by the Federal Statistical Office of Germany within the co-operation project "Machine Learning in Official Statistics". 


\bibliography{literature}

\end{document}


\onecolumn 
\maketitle

\appendix

\section{PSEUDO-CODE FOR BPLS}

We summarize the procedure of Bayesian Pseudo-Label Selection (BPLS) with approximate Pseudo Posterior Predictive (PPP) in Algorithm~\ref{alg:main}. Pseudo-code describing the proposed extensions can be found in section \ref{sec:app-extensions} of this supplementary material. Notation and mathematical symbols follow the main paper. Notably, the number of unobserved data $\lvert \mathcal{U} \rvert$ was denoted $m$ in the main paper.

\RestyleAlgo{ruled}

\SetKwComment{Comment}{/* }{ */}

\begin{algorithm}[H]
\caption{Bayesian Pseudo-Label Selection (BPLS) with approximate Pseudo Posterior Predictive (PPP)}
\label{alg:main}

\KwData{$\mathcal{D}, \mathcal{U}$}
\KwResult{$\mathcal{D}$, fitted model $\hat y^*(x)$}
\textbf{Fit} model M on labeled data $\mathcal{D}$ to obtain prediction function $\hat y(x)$ \\
\While{stopping criterion not met}{
\For{$i \in \{1, \dots, \lvert \mathcal{U} \rvert \}$}{
\textbf{predict} $\mathcal{Y} \ni \hat y_i = \hat y(x_i)$ \\
\textbf{approximate}  PPP $p(\mathcal{D} \cup \left(x_{i}, \hat y_i\right) | \mathcal{D}) $ 
\\
}
\textbf{obtain} $i^* = \argmax_i \{p(\mathcal{D} \cup \left(x_{i}, \hat y_i\right) | \mathcal{D}) \} $ \\ 
\textbf{add} $(x_i, \hat y_i)$ to labeled data: $\mathcal{D} \leftarrow \mathcal{D} \cup (x_i, \hat y_i) $ \\
\textbf{update} $\mathcal{U} \leftarrow \mathcal{U} \setminus \left(x_{i}, \mathcal{Y}\right)_i $

}
\end{algorithm}

\newpage
\section{MISSING PROOFS}

We present the proofs for Theorems 1-3 in section 2 of the main paper. For the sake of readability, we repeat the underlying theorems as well.

\subsection{Proof of Theorem 1}

\begin{theorem}
\label{th:bayes-opt}
In the decision problem $(\mathbb{A}, \Theta, u(\cdot))$ with $\mathbb{A} = \mathcal{U}$ (definition 1), with the pseudo-label likelihood as utility function (definition 2), and a prior $\pi(\theta)$ on $\Theta$, the standard Bayes criterion
\begin{align*}
    \Phi(\cdot,\pi) \colon \mathcal{U} \to \mathbb{R}\\
    a &\mapsto \Phi(a, \pi) = \mathbb{E}_\pi(u(a,\theta)) 
\end{align*}

corresponds to the pseudo marginal likelihood $p(\mathcal{D}~\cup~(x_i, \hat{y}_i))$.
    
\end{theorem}

\begin{proof}
    The definition of the expected value for measurable $u(\cdot, \cdot)$ directly delivers $ \Phi(a, \pi) = \mathbb{E}_\pi(u(a,\theta)) = \int u(a, \theta) d \pi(\theta) = \int p(\mathcal{D} \cup (x_i, \hat{y}_i)\mid\theta) d \pi(\theta) = p(\mathcal{D} \cup (x_i, \hat{y}_i))$.
\end{proof}

\subsection{Proof of Theorem 2}

\begin{theorem}
In the decision problem $(\mathbb{A}, \Theta, u(\cdot))$, using the pseudo-label likelihood as utility function as in theorem \ref{th:bayes-opt} but with the prior updated by the posterior $\pi(\theta) = p(\theta \mid \mathcal{D})$ on $\Theta$, the standard Bayes criterion 
$\Phi(\cdot, \pi) \colon \mathcal{U} \to \mathbb{R}; \, a \mapsto \Phi(a, \pi) = \mathbb{E}_\pi(u(a,\theta)) $
corresponds to the \textit{pseudo posterior predictive} $p(\mathcal{D} \cup (x_i, \hat{y}_i)\mid \mathcal{D})$.
\end{theorem}

\begin{proof}  
    Analogous to Proof 1, we have $ \Phi(a, \pi) = \mathbb{E}_\pi(u(a,\theta)) = \int u(a, \theta) d \pi(\theta).$ Now with the updated prior $\pi(\theta) = p(\theta \mid \mathcal{D})$ it follows $ \int u(a, \theta) d \pi(\theta)= \int p(\mathcal{D} \cup (x_i, \hat{y}_i)\mid\theta) d p(\theta \mid \mathcal{D}) = p(\mathcal{D} \cup (x_i, \hat{y}_i)\mid \mathcal{D})$.
\end{proof}

\subsection{Proof of Theorem 3}

\begin{theorem}
In the decision problem $(\mathbb{A}, \Theta, u(\cdot))$, using the pseudo-label likelihood as utility function as in theorem \ref{th:bayes-opt}, the max-max criterion
\begin{align*}
    \Phi \colon \mathcal{U} \to \mathbb{R}\\
    a &\mapsto \Phi(a) = \max_\theta (u(a,\theta)) 
\end{align*}

corresponds to the (full) likelihood at $\hat \theta_{ML}$.
\end{theorem}

\begin{proof} 
Recall definition 2 of the pseudo-label likelihood as utility function: $ u \colon \mathcal{U} \times \Theta \to \mathbb{R} \; ; \; ((x_i, \mathcal{Y}), \theta) \mapsto u((x_i, \mathcal{Y}), \theta) = p(\mathcal{D} \cup (x_i, \hat{y}_i)\mid \theta).$
Thus, it holds for the max-max criterion $\Phi(a) = \max_\theta (u(a,\theta)) = \max_\theta (p(\mathcal{D} \cup (x_i, \hat{y}_i)\mid \theta)) = p(\mathcal{D} \cup (x_i, \hat{y}_i)\mid \hat \theta_{ML})$, with $\hat \theta_{ML}$ the ML-estimator. 
\end{proof}

The max-max criterion hence corresponds to direct optimization with regard to $a$ of the likelihood, evaluated at $\hat \theta_{ML}$. The respective max-max-action is thus $ a^*_{max-max} = \max_a \max_\theta p(\mathcal{D} \cup (x_i, \hat{y}_i)\mid \theta) = \max_a p(\mathcal{D} \cup (x_i, \hat{y}_i)\mid \hat \theta_{ML})$.

\newpage
\section{EXPERIMENTAL SETUP}

We describe the setup for the experiments with both the simulated and the real-world data along with additional empirical results comparing our approximate PPP with predominant PLS methods in section \ref{sec:add-res}. 

\subsection{Benchmarks}
\label{sec:exp-setup}
Throughout our experiments, we compare our proposed approximate PPP with a set of baseline and competing approaches: 

\begin{itemize}
    \item \textit{Likelihood (max-max)}: Self-training using the Likelihood max-max action as selection criterion 
    \item \textit{Predictive Variance}: Self-training using the predictive variance of the model predictions as a selection criterion 
    \item \textit{Probability Score}: Self-training using the predicted probabilities (scores) as a selection criterion 
    \item \textit{Supervised Learning}: regular supervised model fitting using the labeled training data only
\end{itemize}

All data sets reflect binary classification tasks with a fairly balanced class label distribution.
Hence, we report and compare with model performance as measured in accuracy on the holdout test data sets.

\subsubsection{Generalized Linear Models}

We choose generalized linear models (GLMs) \cite{nelder1972generalized} as predictive models for BPLS with PPP as well as for all competing methods listed in section \ref{sec:exp-setup}. By considering the binomial distribution from the exponential family this yields logistic regression:

\begin{equation}
  \mathrm {P} (Y=1\mid X=x_{i})=\mathrm {P} (Y_{i}=1)={\frac {\exp(\mathbf {x} _{i}^{\top }{\boldsymbol {\beta }})}{1+\exp(\mathbf {x} _{i}^{\top }{\boldsymbol {\beta }})}}={\frac {1}{1+\exp(-\mathbf {x} _{i}^{\top }{\boldsymbol {\beta }})}},
\end{equation}

with $\boldsymbol{\beta }=(\beta _{0},\beta _{1},\ldots ,\beta _{k})^{\top }$ and $\mathbf {x} _{i}^{\top }{\boldsymbol {\beta }}=\beta _{0}+x_{i1}\beta _{1}+x_{i2}\beta _{2}+\dotsc +x_{ik}\beta _{k}$. Such a regression with additive linear predictor $\mathbf {x} _{i}^{\top }{\boldsymbol {\beta }}$ can be easily be extended to target variables that follow a multinomial distribution (i.e. multi-class problems). Our setup described in section \ref{sec:exp-setup} can thus be extended in a straightforward manner to such learners for multi-class classification tasks.

\subsubsection{Generalized Additive Models}
 
We also use non-parametric generalized additive models (GAMs) \cite{fahrmeir2013regression, hastie2017generalized} as predictive models. Here, the response variable depends on unknown smooth functions of some feature variables: 

\begin{equation}
    g(\mathbb{E}(Y))=\beta _{0}+f_{1}(x_{1})+f_{2}(x_{2})+\cdots +f_{m}(x_{m}).
\end{equation}

As above, we assume $Y$ to follow a binomial distribution in our experiments, since we only consider binary classification. Like GLMs, GAMs can be easily extended to multi-class problems.

\subsubsection{Simulation Design}

For the simulation study, we created a simulated dataset with $n$ samples for a binary classification based on a varying amount of $q$ features. This simulation follows the model equation

\begin{equation}
    y_i \sim Bin(1, p_i), \;
    \text{with} \; p_i = \left(1 + exp(- x_{i,0} + x_{i,1} + ... x_{i,p})\right)^{-1}
\end{equation}

where $x_i \sim \mathcal{N}(\mu, \sigma^2)$ independently with varying $\mu$ and $\sigma^2$.

\subsubsection{Pre-Processing and Gathering of Real-world Data}

Detailed information on sources, features, and target variables of all data sets \cite{Dua:2019} that were used in the experiments can be found in section \ref{sec:data-sets}. The data sets were selected randomly after filtering according to the following criteria:
\begin{itemize}
    \item We only consider binary classification tasks, since we test the PLS methods based on semi-supervised logistic regression.
    \item We choose from datasets with a low number of missing values in order to minimize algorithm differences in missing value handling.
    \item We restrict ourselves to datasets with $q < 100$ to avoid massive overfitting and computational trouble.
\end{itemize}

In order to benchmark BPLS against classical PLS methods, we split the data sets into train and test data first, before removing labels from a pre-defined share of training data. Our detailed splitting procedure for the real-world datasets with a total size of $n$ samples each is the following:

\begin{enumerate}
    \item draw $n_{test}$ samples to create the holdout test set $D_{test}$ where the remainder constitutes the training set $D_{train}$ of size $n_{train}$ such that $n_{train} = n_{test}$  (share of test data thus $50 \%$). 
    \item draw $n_{labeled}$ samples from $D_{train}$ to create the labeled training data $D_{train}^{labeled}$ 
    \item Remove labels from remaining samples in $D_{train}$ and treat them as unlabeled data $D_{train}^{unlabeled}$
\end{enumerate}

Throughout our experiments, we repeat self-training $R$ times and use varying shares of labeled data $\frac{n_{unlabeled}}{n_{train}}$.

\subsubsection{Hypotheses}

For interpretation purposes, recall our hypotheses that we specified before running the experiments:

\begin{hypothesis}
\label{hypo:BPLS-good}
\textbf{(a)} PPP with uninformative prior outperforms traditional PLS on data prone to initial overfitting (i.e., with high ratio of features to data $\frac{p}{n}$ and poor initial generalization).   
\textbf{(b)} For low $\frac{p}{n}$ and high initial generalization, BPLS is outperformed by traditional PLS.
\end{hypothesis}

\begin{hypothesis}
\label{hypo:likelihood}
    \textbf{(a)} Among all PLS methods, the pseudo-label likelihood (max-max-action) reinforces the initial model fit the most and \textbf{(b)} hardly improves generalization. 
\end{hypothesis}

\begin{hypothesis}
\label{hypo:informative}
    PPP with informative prior outperforms traditional PLS methods universally.
\end{hypothesis}

\newpage

\section{FURTHER RESULTS}
\label{sec:add-res}

In this section, we present additional results. 
Section \ref{sec:res-sim} has the complete results for simulated data with $q = 60$ features.\footnote{Results for $n = 100$ and $n = 400$ were already included in the paper, but are also shown here for the sake of completeness of the setup with $q = 60$. (Note that this is an exception; all other results presented herein have not been included in the paper.)} In section \ref{sec:res-sim-further}, we show additional results for smaller $q \in \{10,15,20,30\}$ with varying $n \in \{300, 400, 800, 1000\}$.

\subsection{Results on Simulated Data with $q = 60$}
\label{sec:res-sim}

\begin{figure}[H]
    \centering
    \includegraphics[width=\textwidth]{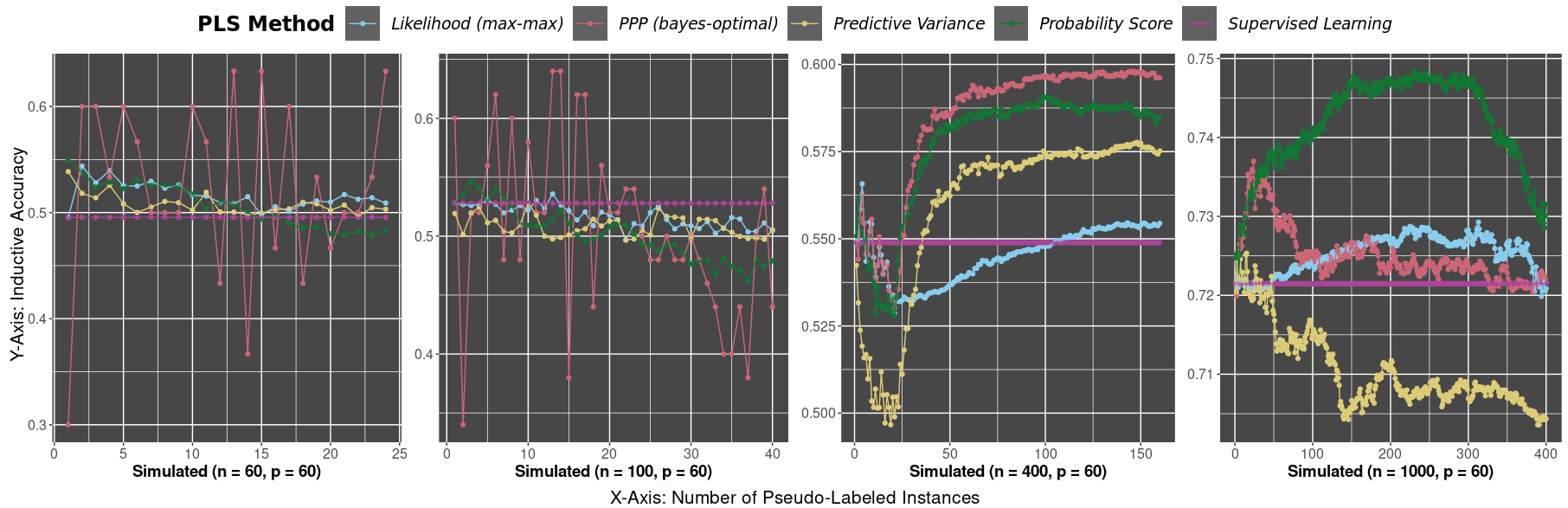}
    \caption{Complete Results on Simulated Data for $q = 60$. $R = 100$; $\frac{n_{unlabeled}}{n_{train}} = 0.8$.}
    \label{fig:my_label}
\end{figure}

\subsection{Further Results on Simulated Data with $q \in \{10,15,20,30\}$}
\label{sec:res-sim-further}

\begin{figure}[H]
    \centering
    \includegraphics[width=\textwidth]{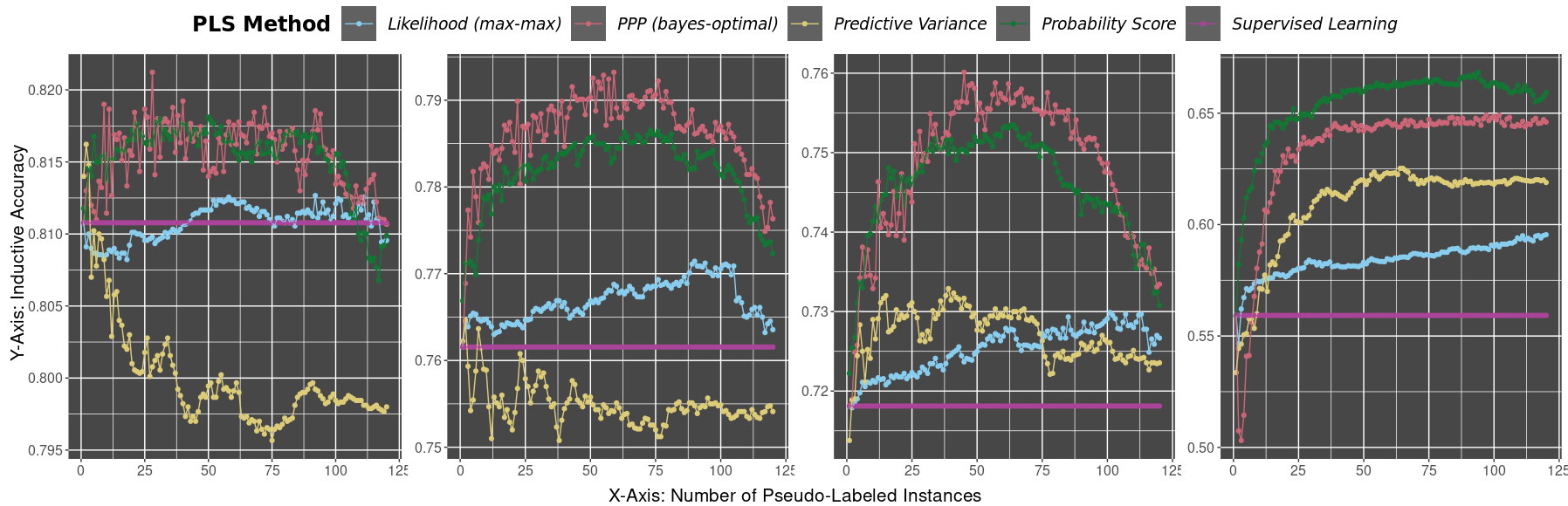}
    \caption{Results on Simulated Data, $n = 300$ and (from left to right) $q \in \{10,15,20,30\}$. $R = 100$; $\frac{n_{unlabeled}}{n_{train}} = 0.8$.}
    \label{fig:my_label}
\end{figure}

\begin{figure}[H]
    \centering
    \includegraphics[width=\textwidth]{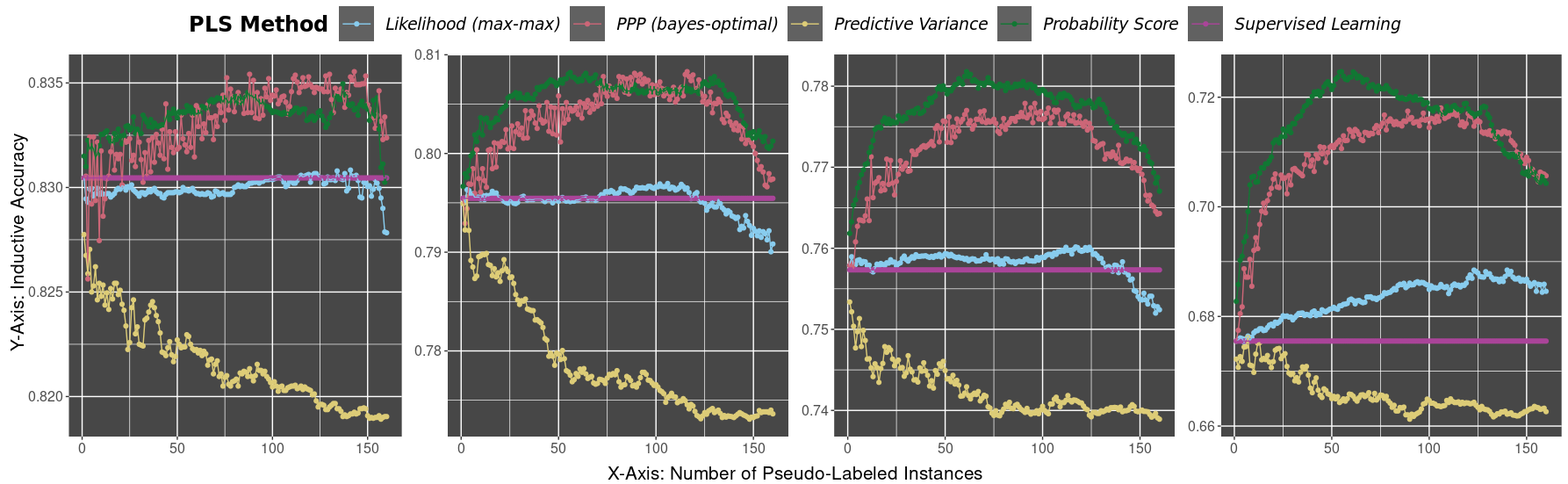}
    \caption{ Results on Simulated Data, $n = 400$ and (from left to right) $q \in \{10,15,20,30\}$. $R = 100$; $\frac{n_{unlabeled}}{n_{train}} = 0.8$.}
    \label{fig:my_label}
\end{figure}

\begin{figure}[H]
    \centering
    \includegraphics[width=\textwidth]{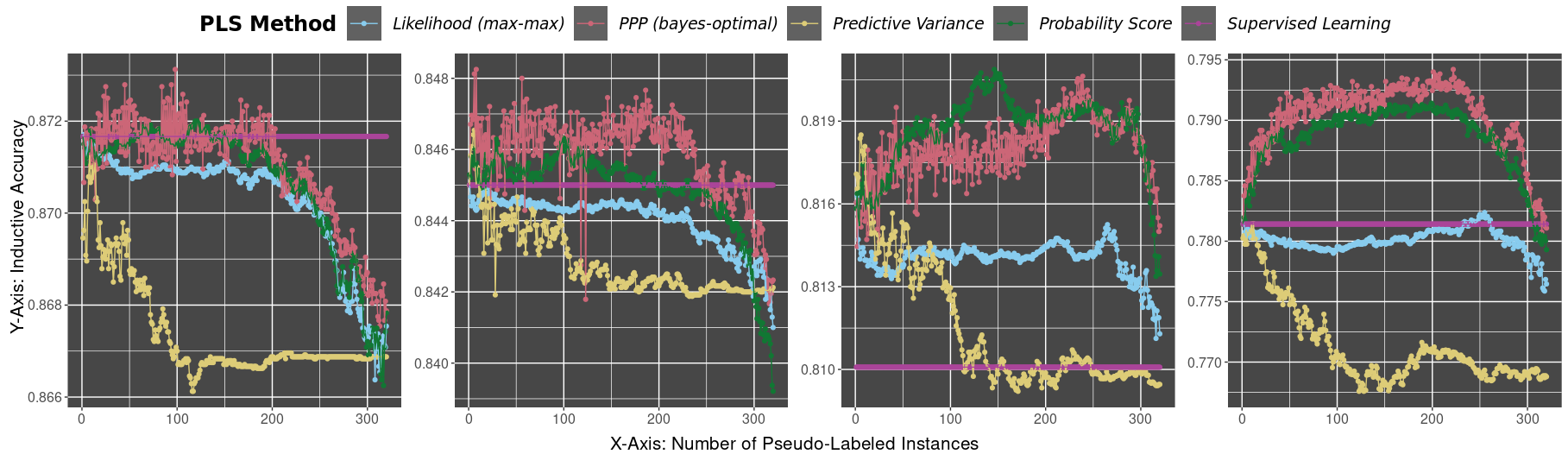}
    \caption{ Results on Simulated Data, $n = 800$ and (from left to right) $q \in \{10,15,20,30\}$. $R = 100$; $\frac{n_{unlabeled}}{n_{train}} = 0.8$.}
    \label{fig:my_label}
\end{figure}

\begin{figure}[H]
    \centering
    \includegraphics[width=\textwidth]{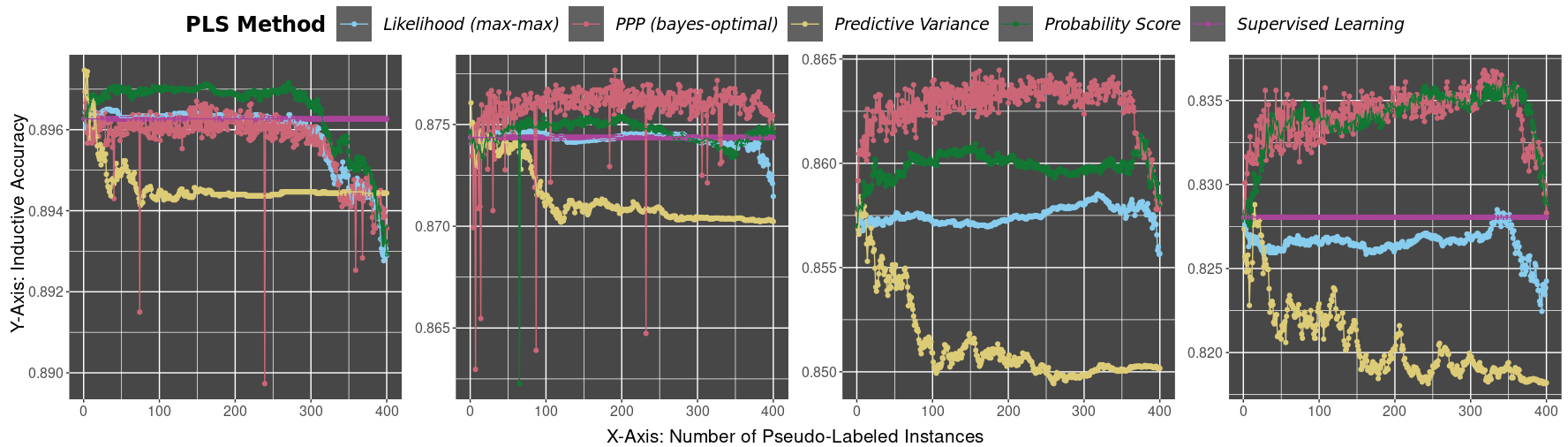}
    \caption{ Results on Simulated Data, $n = 1000$ and (from left to right) $q \in \{10,15,20,30\}$. $R = 100$; $\frac{n_{unlabeled}}{n_{train}} = 0.8$.}
    \label{fig:my_label}
\end{figure}

\newpage

\begin{landscape}
\thispagestyle{empty}

\subsection{Informative Prior: Further Results on Simulated Data}

\begin{figure*}[h!]
\centering
\begin{minipage}[b]{0.99\linewidth}
\centering
\includegraphics[scale = 0.23]{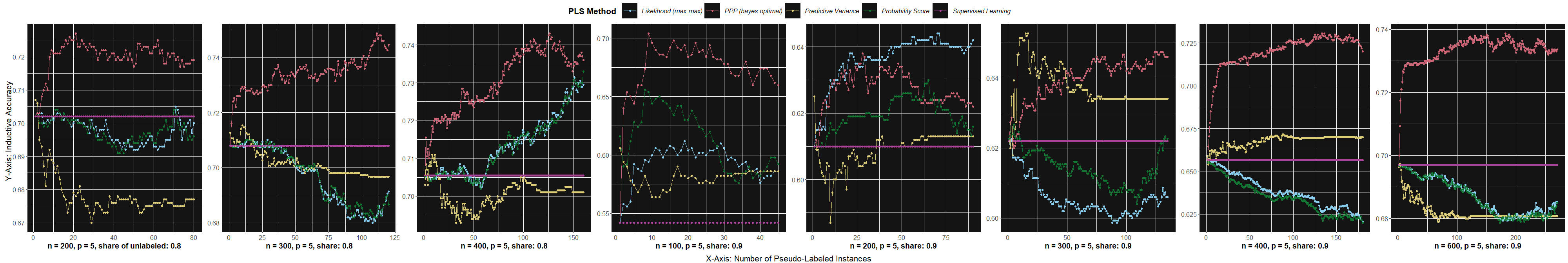}
\end{minipage}
\begin{minipage}[b]{0.99\linewidth}
\centering
\includegraphics[scale = 0.23]{figures/Rplot001(1).png}
\end{minipage}
\caption{Results from simulated data in case of informative priors with simple GLMs (logistic regression, first row) and more complex non-parametric GAMs (second row). Note that resolution allows zooming in.}
\label{fig:res:sim-inf}
\end{figure*}

\vfill
\raisebox{}{\makebox[\linewidth]{\thepage}}

\end{landscape}

\newpage

\subsection{Summary of Results on Simulated Data}
Table \ref{tab:table} summarizes the results on simulated data in an ordinal manner. That is, it shows the best-performing method on the different setups. As in the main paper, \say{Oracle stopping} in table \ref{tab:table} refers to comparing PLS methods with regard to their overall best accuracy as opposed to \say{final} comparisons after the whole data set was labeled.

\begin{table}[H]
\caption{Best performing PLS on Simulated Data} 
\begin{center}
\begin{tabular}{c||c||ll}
\textbf{n} & \textbf{p} &\textbf{ORACLE STOPPING} & \textbf{FINAL} \\
\hline \hline
60 & 60 &  PPP         & PPP\\
100 & 60 & PPP & Supervised Learning \\
400 & 60 & PPP             & PPP \\
1000 & 60 & Probability Score & Probability Score \\
\hline
300 & 30 & Probability Score & Probability Score \\
300 & 20 & PPP & PPP \\
300 & 15 & PPP & PPP \\
300 & 10 & PPP/Probability Score & PPP/Probability Score \\
\hline
400 & 30 & Probability Score & PPP/Probability Score \\
400 & 20 & Probability Score & Probability Score \\
400 & 15 & PPP/Probability Score & Probability Score \\
400 & 10 & PPP/Probability Score & PPP/Probability Score \\
\hline
800 & 30 & PPP & PPP \\
800 & 20 & PPP/Probability Score & PPP/Probability Score \\
800 & 15 & PPP & PPP \\
800 & 10 & PPP & PPP/Probability Score \\
\hline
1000 & 30 & PPP & PPP/Probability Score  \\
1000 & 20 & PPP & PPP \\
1000 & 15 & PPP & PPP \\
1000 & 10 & Predictive Variance & Predictive Variance \\
\label{tab:table}

\end{tabular}
\end{center}
\end{table}

\newpage

\newpage

\section{EXTENSIONS}
\label{sec:app-extensions}

We provide further details on the suggested extensions in section 6. Besides, we briefly discuss other potential extensions.

\subsection{Extensions proposed in the Paper}

We summarize the proposed extensions' procedure from section 6 in the paper by pseudo-code as follows.

\subsubsection{Bivariate Pseudo-Label Selection}

The idea of bivariate BPLS would be to touch the model class $M$. When comparing PPPs, one could then take into account the required model size $q$. The rough idea would be to prefer pseudo-labels that have high plausibility (high likelihood) even with simpler models (small $q$).

\RestyleAlgo{ruled}

\SetKwComment{Comment}{/* }{ */}

\begin{algorithm}[H]
\caption{Bivariate Bayesian Pseudo-Label Selection (BPLS)}

\KwData{$\mathcal{D}, \mathcal{U}$}
\KwResult{$\mathcal{D}$, fitted model $\hat y^*(x)$}
\textbf{Fit} model M on labeled data $\mathcal{D}$ to obtain prediction function $\hat y(x)$ \\
\While{stopping criterion not met}{
\For{$i \in \{1, \dots, \lvert \mathcal{U} \rvert \}$}{
\textbf{predict} $\mathcal{Y} \ni \hat y_i = \hat y(x_i)$ with models of varying $dim(\Theta)$ \\
\textbf{evaluate} PPP $p(\mathcal{D} \cup \left(x_{i}, \hat y_i\right) | \mathcal{D}) $ with predictions from the best performing (on training data) model and save respective $dim(\Theta)$  
\\
}
\textbf{obtain} $i^* = \argmax_i \{f(p(\mathcal{D} \cup \left(x_{i}, \hat y_i\right) | \mathcal{D}), dim(\Theta)) \} $, with $f(\cdot, \cdot)$ some linear combination of the PPP and the model size $dim(\Theta)$ \\ 
 \textbf{retrain} M on $\mathcal{D} \cup \left(x_{i}, \hat y_{i^*}\right)$ \\
\textbf{predict} $\mathcal{Y} \ni \hat y_i^*(\textbf{x} \cup x_i), \textbf{x} \in \mathcal{D} $ \\
\textbf{add} $(x_i, \hat y_i)$ to labeled data: $\mathcal{D} \leftarrow \mathcal{D} \cup (x_i, \hat y_i) $ \\
\textbf{update} $\mathcal{U} \leftarrow \mathcal{U} \setminus \left(x_{i}, \mathcal{Y}\right)_i $

}
\end{algorithm}

\subsection{Additional Extensions}

\subsubsection{Robust PPP}

We further propose a robust extension of PPP based on generalized Bayesian inference \cite{dempster1968generalization, walley1991statistical, insua2012robust, augustin2014introduction}. Recall that for the \textit{robust} PPP, now denoted as $p^*(\hat{y} \mid x, \bm y, \bm x)$, we consider the prior $\pi^*(\theta )$ among all priors from a convex set of priors $\Pi$ that has the smallest value $\pi^*(\hat \theta )$ at the ML-estimator $\hat \theta$. Recall that $\Pi \subseteq \{\pi(\theta) \mid \pi(\cdot) \, \text{a probabilty measure on } \left(\Theta, \sigma(\Theta) \right) \}$ with $\Theta$ compact as throughout the paper and $\sigma(\cdot)$ an appropriate $\sigma$-algebra.

More formally and encapsulating the notion of $\Gamma$-Maximin as in \cite{guo2010decision}, for instance, we have the decision problem $(\mathbb{A}, \Theta, u(\cdot))$ with $\mathbb{A} = \mathcal{U}$ (definition 1 in paper) with the pseudo-label likelihood as utility function (definition 2) and a set of priors $\Pi$ as above. Then the $\Gamma$-maximin criterion 

\begin{equation}
\Phi(\cdot,\Pi) \colon \mathcal{U} \to \mathbb{R}; \,
a \mapsto \Phi(a, \pi) = \underline{\mathbb{E}}_\Pi(u(a,\theta))
\end{equation}

with $\underline{\mathbb{E}}_\Pi(u(a,\theta)) = \inf_{\pi \in \Pi} \mathbb{E}(u(a,\theta))$ corresponds to the \textit{robust pseudo posterior predictive} $p^*(\mathcal{D} \cup (x_i, \hat{y}_i)\mid \mathcal{D})$ that results from updating the prior $\pi^*(\cdot) \in \Pi$ that has the lowest value in $\hat \theta$. Action $a_{\Gamma}^* = \argmax_i p^*(\mathcal{D} \cup (x_i, \hat{y}_i)\mid \mathcal{D})$ is $\Gamma$-maximin-optimal for prior $\pi^*(\cdot)$.

In practice, the proposed extension heavily depends on the exact nature of $\Pi$. For illustrative purposes, suppose that we can specify $\Pi$ such that the most contradicting prior is such that the resulting posterior is uniform. Effectively, we then end up with the same situation as with the marginal likelihood when the prior is uniform in case of independent observations, see the end of section 3.1 in the main paper: We randomly select pseudo-labeled instances. Quite intuitively, the selection that is most robust toward the initial fit given no other information is just such a random selection.

\subsubsection{Bayesian Pseudo-Label Selection without predictions}
\label{sec:no-preds}
The idea here would be to directly assign all possible $q$ classes in $\mathcal{Y}$ to the unlabeled data points with $q = \lvert \mathcal{Y} \rvert$. The following pseudo-code lines out the procedure. Note that the inner loop thus requires $\lvert \mathcal{U} \rvert \cdot \lvert \mathcal{Y}\rvert$ assignments and respective PPP evaluations.

\RestyleAlgo{ruled}

\SetKwComment{Comment}{/* }{ */}

\begin{algorithm}[H]
\caption{Bayesian Pseudo-Label Selection (BPLS) without predictions}

\KwData{$\mathcal{D}, \mathcal{U}$}
\KwResult{$\mathcal{D}$, fitted model $\hat y^*(x)$}
\textbf{Fit} model M on labeled data $\mathcal{D}$ to obtain prediction function $\hat y(x)$ \\
\While{stopping criterion not met}{
\For{$i \in \{1, \dots, \lvert \mathcal{U} \rvert \}$}{
\textbf{assign}  all possible $\hat y_i \in \mathcal{Y}$ to $\left(x_{i}, \hat y_i\right)$ \\
\textbf{evaluate} all possible PPP $p(\mathcal{D} \cup \left(x_{i}, \hat y_i\right) | \mathcal{D}) $ 
\\
}
\textbf{obtain} $i^* = \argmax_i \{p(\mathcal{D} \cup \left(x_{i}, \hat y_i\right) | \mathcal{D}) \} $ \\ 
 \textbf{retrain} M on $\mathcal{D} \cup \left(x_{i}, \hat y_{i^*}\right)$ \\
\textbf{predict} $\mathcal{Y} \ni \hat y_i^*(\textbf{x} \cup x_i), \textbf{x} \in \mathcal{D} $ \\
\textbf{add} $(x_i, \hat y_i)$ to labeled data: $\mathcal{D} \leftarrow \mathcal{D} \cup (x_i, \hat y_i) $ \\
\textbf{update} $\mathcal{U} \leftarrow \mathcal{U} \setminus \left(x_{i}, \mathcal{Y}\right)_i $

}
\end{algorithm}

\subsubsection{Fantasy PPP}

In complete analogy to the proposed extension in section \ref{sec:no-preds}, we consider assignment of all possible classes instead of predictions of single classes. As opposed to selecting from all possible pseudo-labels directly, we could also combine the PPPs from pseudo-labels for each instance to a fantasy PPP by a weighted sum. See the following pseudo-code for details. The formulation allows for different ways of how to define the weighted sum $\Sigma$. Regarding one instance, we would have a PPP for each class $y \in \mathcal{Y}$. One way to define $\Sigma$ would be to consider the maximal and minimal PPP only and compute a weighted sum thereof, leaning on the Hurwicz-criterion in decision theory \cite{hurwicz1951generalized}. The weight assigned to the maximal PPP is then regarded the decision-maker's degree of optimism.

\begin{algorithm}[H]
\caption{Bayesian Pseudo-Label Selection (BPLS) with fantasy PPPs}

\KwData{$\mathcal{D}, \mathcal{U}$}
\KwResult{$\mathcal{D}$, fitted model $\hat y^*(x)$}
\textbf{Fit} model M on labeled data $\mathcal{D}$ to obtain prediction function $\hat y(x)$ \\
\While{stopping criterion not met}{
\For{$i \in \{1, \dots, \lvert \mathcal{U} \rvert \}$}{
\textbf{assign} all possible $y_i \in \mathcal{Y}$ to $\left(x_{i}, y_i\right)_i$ \\
\textbf{evaluate} weighted sum $\Sigma$ of respective PPPs $p(\mathcal{D} \cup \left(x_{i}, y_i\right) | \mathcal{D}) $ 
\\
}
\textbf{obtain} $i^* = \argmax_i \Sigma $ \\ 
 \textbf{retrain} M on $\mathcal{D} \cup \left(x_{i}, \hat y_{i^*}\right)$ \\
\textbf{predict} $\mathcal{Y} \ni \hat y_i^*(\textbf{x} \cup x_i), \textbf{x} \in \mathcal{D} $ \\
\textbf{add} $(x_i, \hat y_i)$ to labeled data: $\mathcal{D} \leftarrow \mathcal{D} \cup (x_i, \hat y_i) $ \\
\textbf{update} $\mathcal{U} \leftarrow \mathcal{U} \setminus \left(x_{i}, \mathcal{Y}\right)_i $

}
\end{algorithm}

\section{Numerical Experiments verifying the Simplified Approximation}

\subsection{Simplified Approximation}

We test the equivalence of PLS with regard to the approximate PPP criterion (Equation (6) in main paper)

\begin{align*} 
  \tilde \ell(\tilde \theta) - \frac 1 2 \log |\mathcal I(\tilde \theta)| + \log \pi(\tilde \theta)
\end{align*}

with $\tilde \ell(\tilde \theta) = \ell_{\Dcal\cup (x_i, \hat y_i)}(\tilde \theta)  + \ell_{\Dcal}(\tilde \theta)$, and our simplified version thereof (Equation (7) in main paper):

\begin{align*} 
   \ell_{\Dcal \cup (x_i, \hat y_i)}(\tilde \theta) - \frac 1 2 \log | \mathcal I(\tilde \theta)| + \log \pi(\tilde \theta).
\end{align*}

Recall that these terms are approximately equivalent when comparing pseudo-samples $(x_i, \hat y_i)$ and $(x_j, \hat y_j)$. We expanded $\ell_{\Dcal}$ around its maximizer $\hat \theta$, so that $\ell_{\Dcal}(\tilde \theta) =  \ell_{\Dcal}(\hat \theta) + O(\|\hat \theta - \tilde \theta\|^2)$. Since  $\Dcal \cup (x_i, \hat y_i)$ and $\Dcal$ differ in only one sample, the difference $\hat \theta - \tilde \theta$ is of order $O(n^{-1})$. Thus,$$ \tilde \ell(\theta) = \ell_{\Dcal\cup (x_i, \hat y_i)}(\theta) + \ell_{\Dcal}(\hat \theta) + O(n^{-2}).$$
The remainder is negligible compared to the other terms in Equation (6) and $\ell_{\Dcal}(\hat \theta)$ does not depend on the pseudo-sample $(x_i, \hat y_i)$. This suggests the simplified \emph{informative BPLS criterion} : $   \ell_{\Dcal \cup (x_i, \hat y_i)}(\tilde \theta) - \frac 1 2 \log | \mathcal I(\tilde \theta)| + \log \pi(\tilde \theta).$

\subsection{Experimental Setup}

In addition to this theoretical argument, we provide empirical evidence for this equivalence. It is verified numerically for small $n$ by experiments on the ionosphere data \cite{ionosphere}, EEG data \cite{zhang1995event}, banknote data \cite{Dua:2019}, abalone data \cite{waugh1995extending} as well as on simulated binomially distributed data, see section \ref{sec:exp-setup}. For all data sets, we compare semi-supervised GLM performance of BPLS with simplified criterion (\say{rough PPP}, eq. 7) and unsimplified criterion (\say{fine PPP}, eq. 6) with regard to test accuracy averaged over 40 repetitions. 

\subsection{Results}

\begin{figure}[!h]
    \centering
    \begin{minipage}{.4\textwidth}
        \centering
        \includegraphics[scale=0.4, trim={0 0.8cm 5cm 0},clip]{Sample UAI 2023 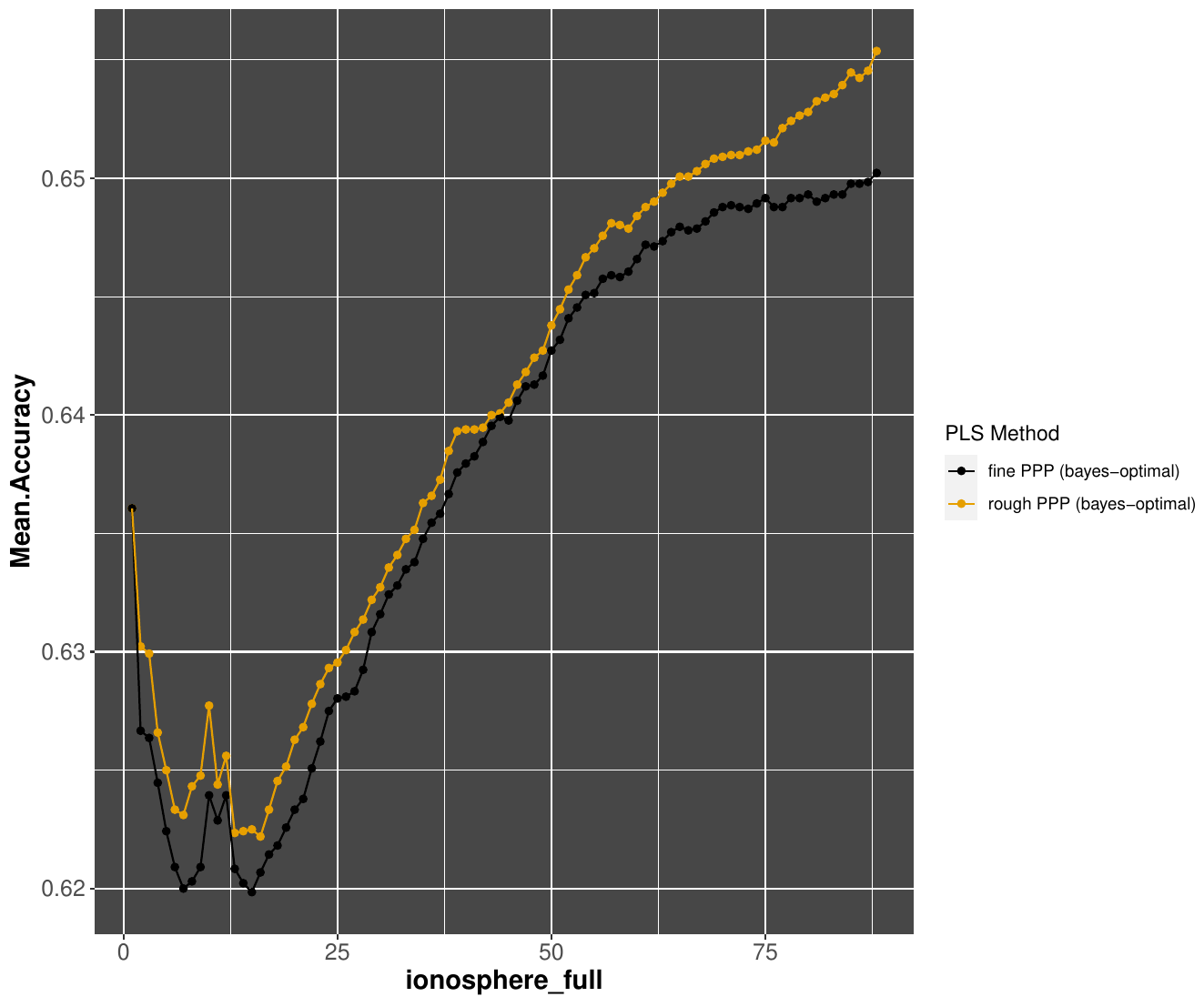}
        \caption{Approximations' performances \\ on ionosphere subsample of size $n=220$.}
        \label{fig:iono-1}
    \end{minipage}%
    \begin{minipage}{0.4\textwidth}
        \centering
        \includegraphics[scale=0.4, trim={0 0.8cm 0cm 0},clip]{Sample UAI 2023 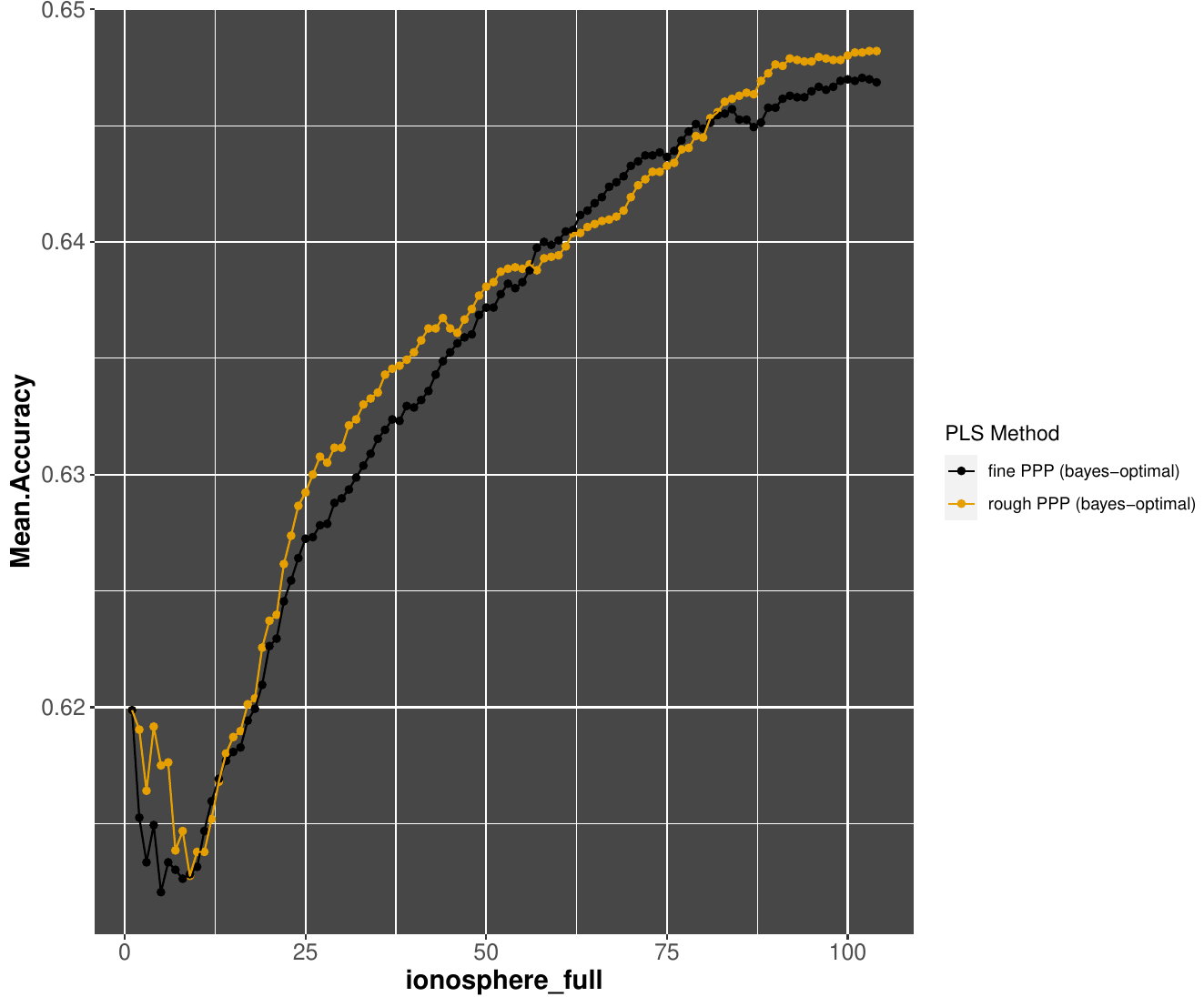}
        \caption{Approximations' performances \\ on ionosphere subsample of size $n=260$.}
        \label{fig:iono-2}
    \end{minipage}
\end{figure}

\begin{figure}[!h]
    \centering
    \begin{minipage}{.4\textwidth}
        \centering
        \includegraphics[scale=0.4, trim={0 0.8cm 5cm 0},clip]{Sample UAI 2023 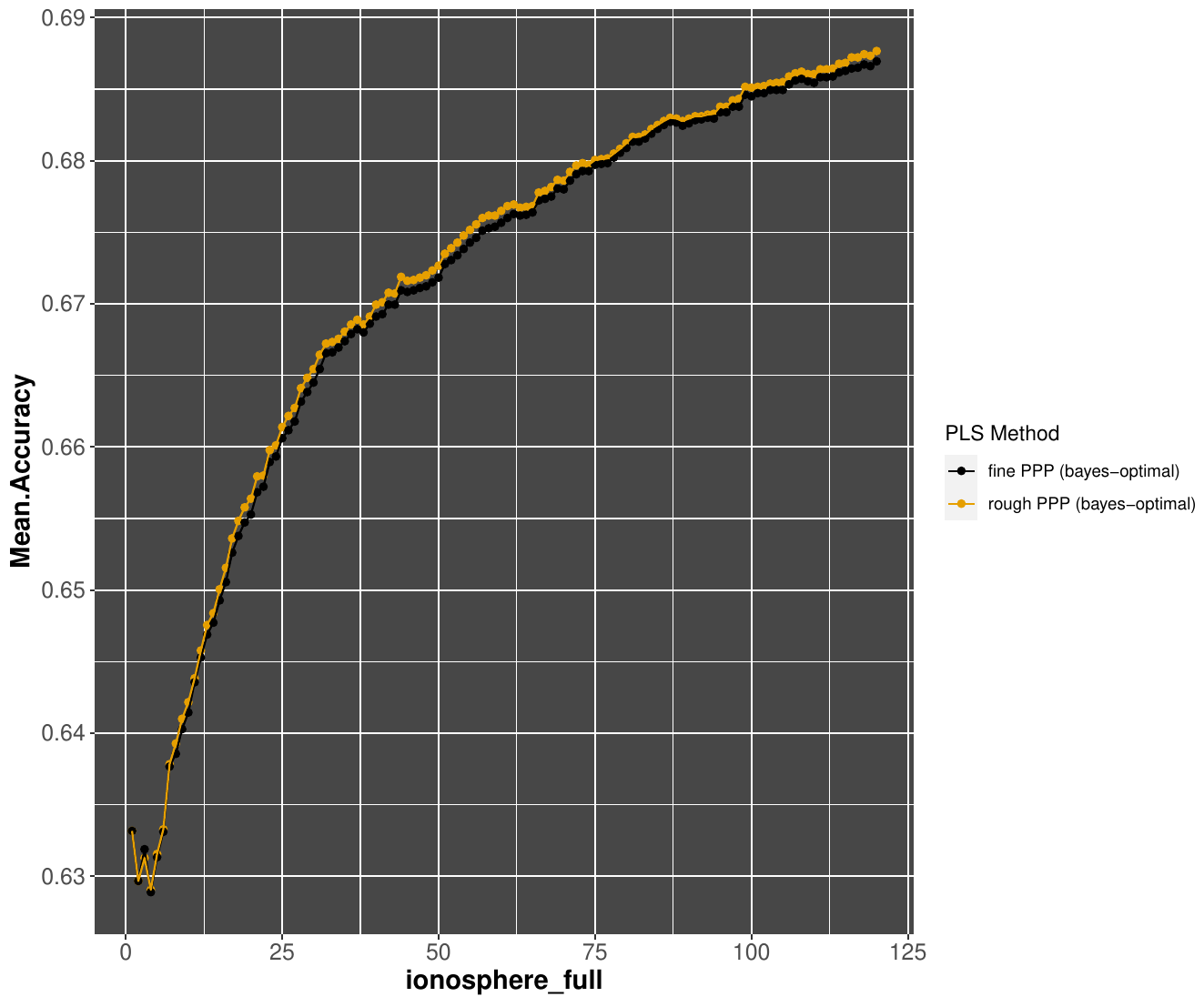}
        \caption{Approximations' performances \\ on ionosphere subsample of size $n=300$.}        \label{fig:iono-3}
    \end{minipage}%
    \begin{minipage}{0.4\textwidth}
        \centering
        \includegraphics[scale=0.4, trim={0 0.8cm 0cm 0},clip]{Sample UAI 2023 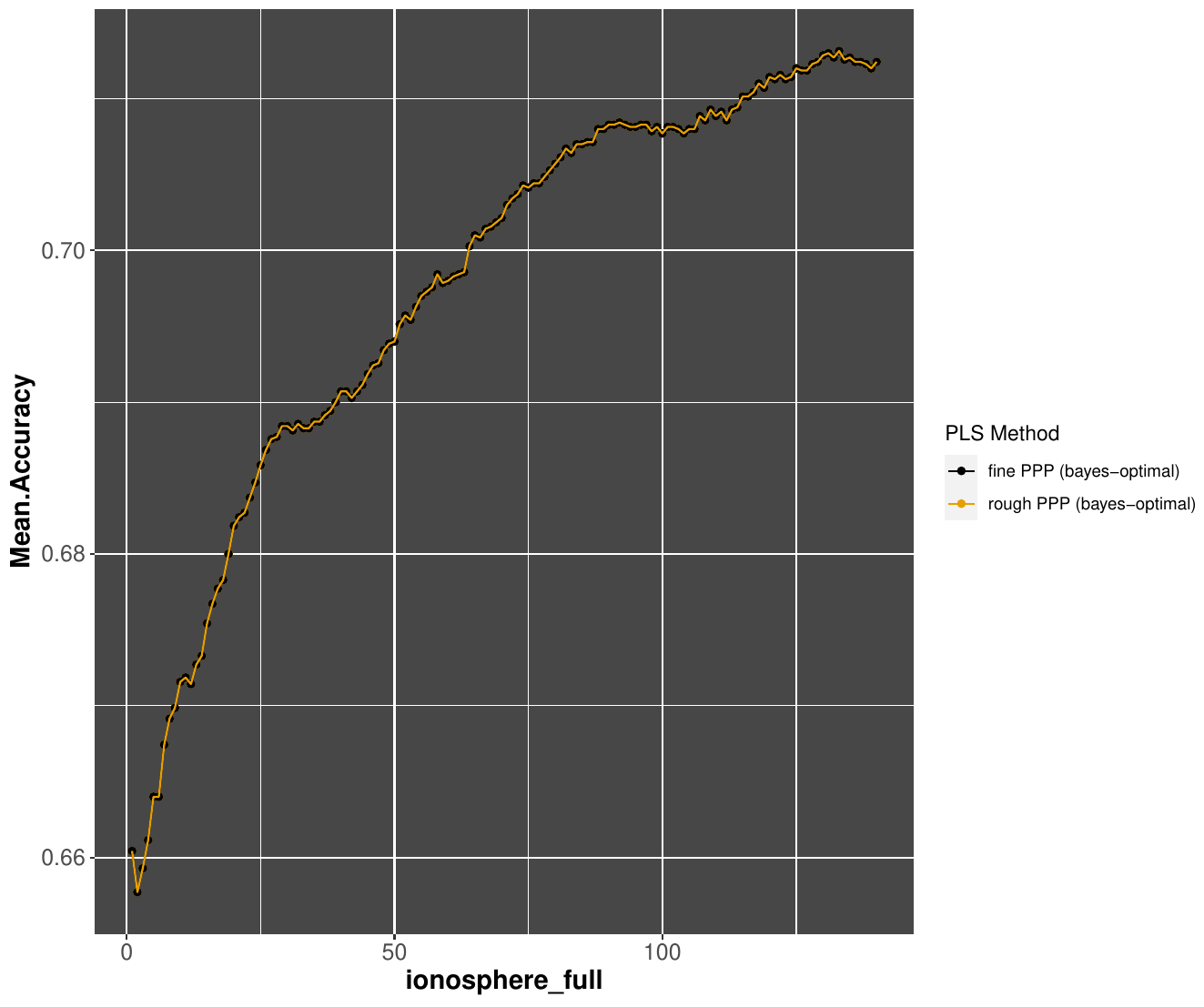}
        
        \caption{Approximations' performances \\ on ionosphere data set of size $n=350$.}
        \label{fig:iono-4}
    \end{minipage}
\end{figure}

\begin{figure}[!h]
    \centering
    \includegraphics[scale=0.7]{Sample UAI 2023 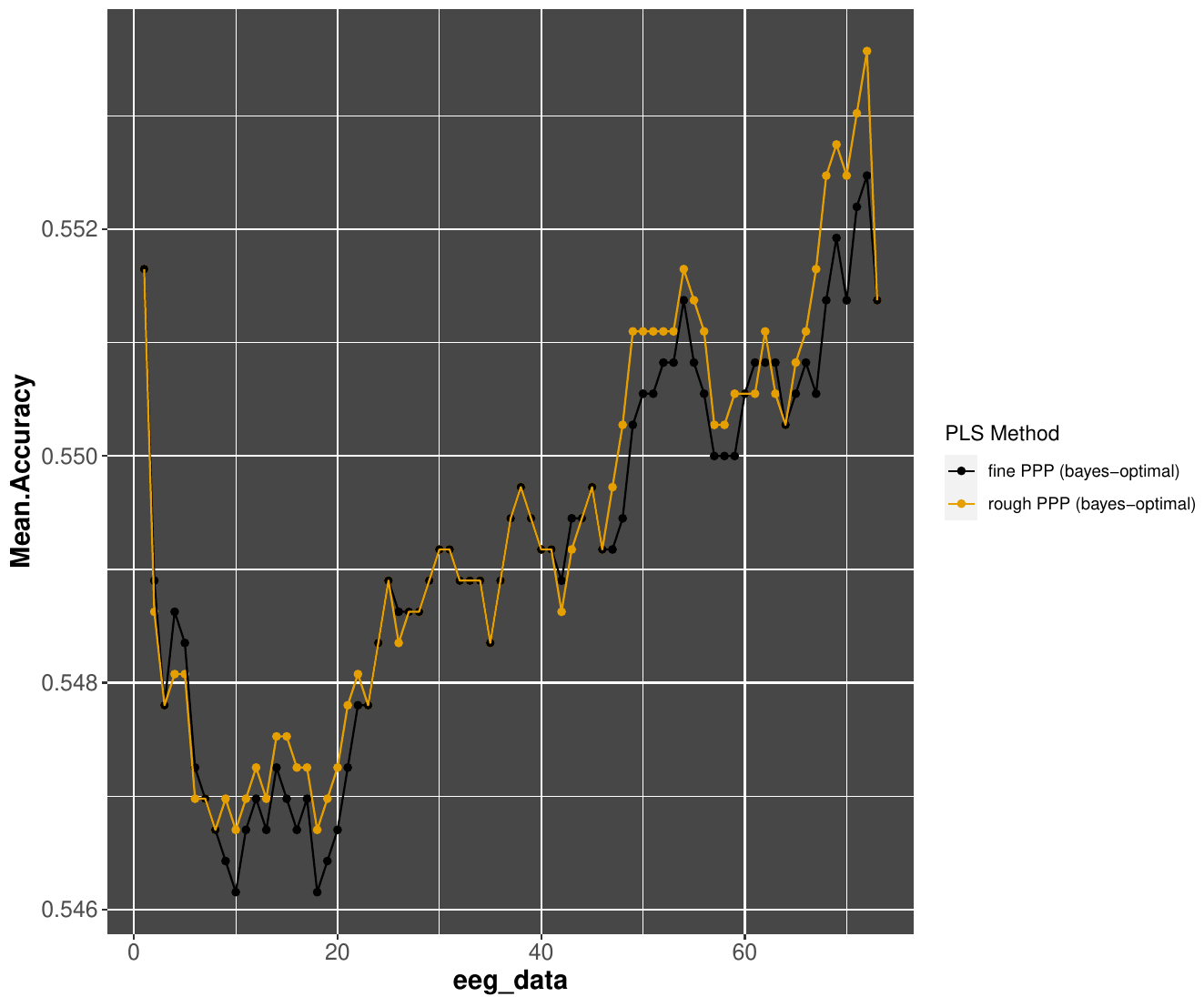}
    \caption{Approximations' performances on EEG data set ($n=185$, $q = 13$).}
    \label{fig:eeg}
\end{figure}

\begin{figure}[!h]
    \centering
    \includegraphics[scale=0.7]{Sample UAI 2023 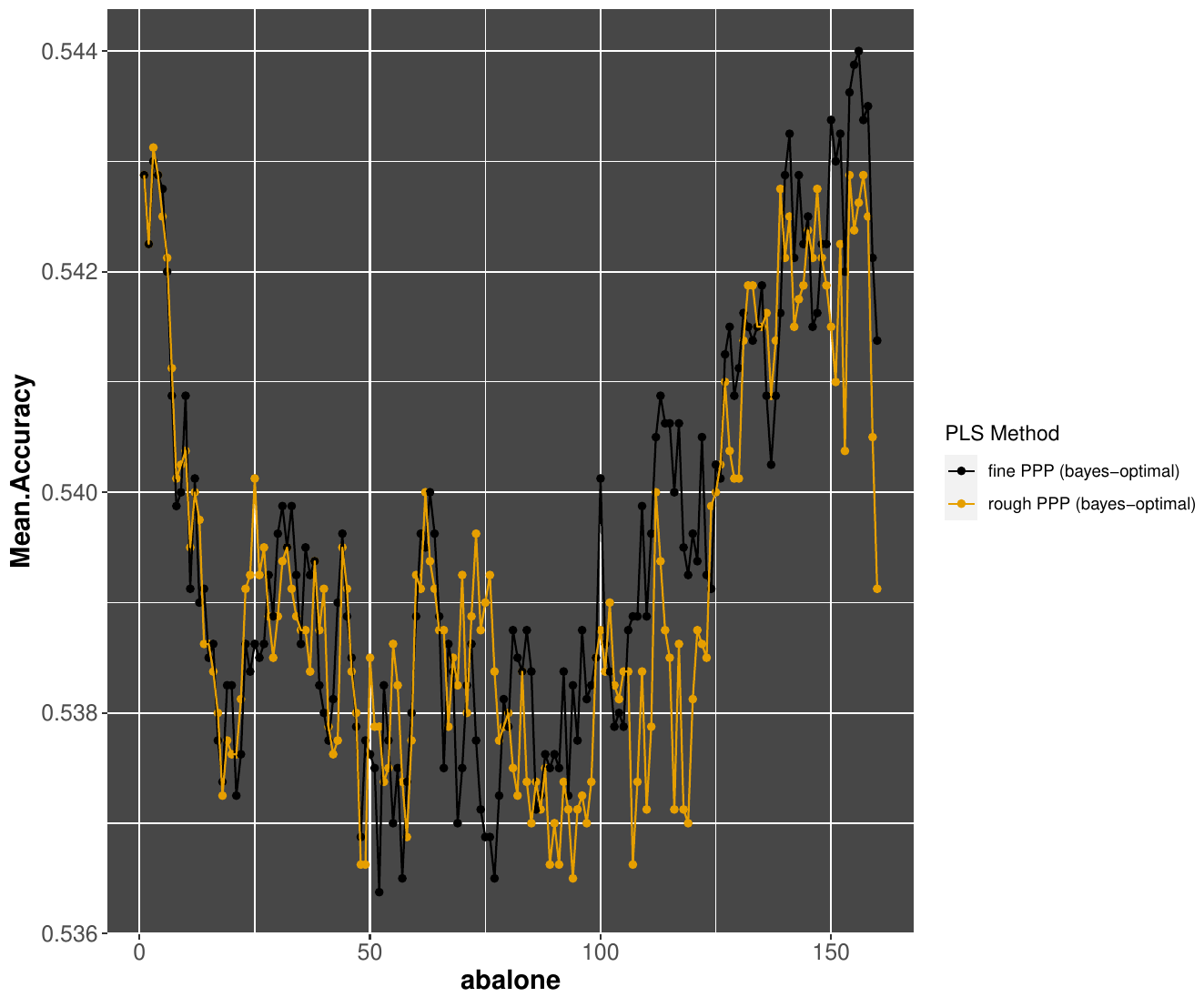}
    \caption{Approximations' performances on abalone data set ($n=400$, $q = 4$).}    \label{fig:abalone}
\end{figure}

\begin{figure}[!h]
    \centering
    \includegraphics[scale=0.7]{Sample UAI 2023 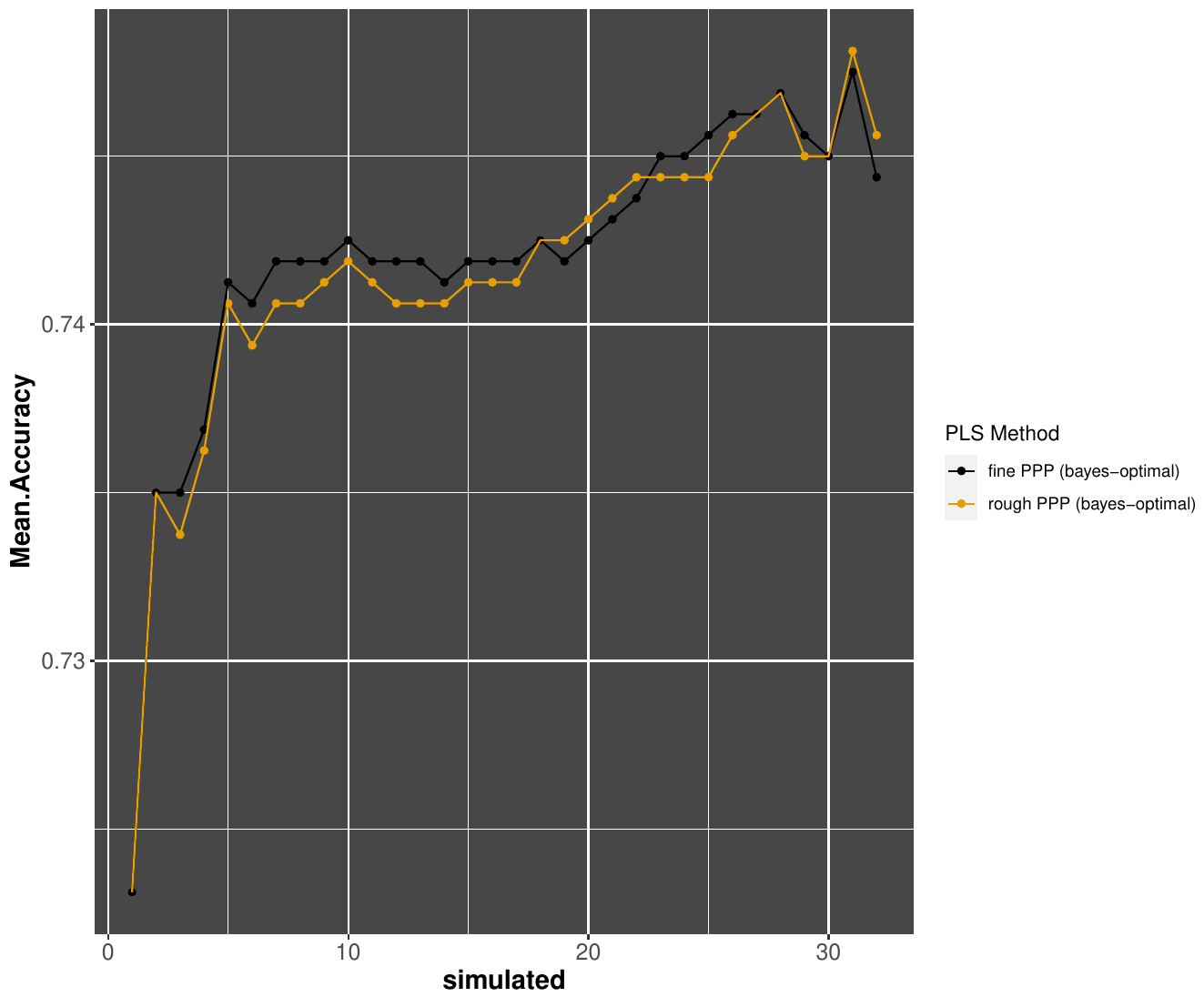}
    \caption{Approximations' performances on simultated data set ($n=120$, $q = 4$).}    \label{fig:simulated}
\end{figure}

Figure~\ref{fig:eeg} shows the results for EEG data, Figure~\ref{fig:abalone} for abalone data, while Figure~\ref{fig:simulated} displayes results for the simulated binomially distributed data. In order to assess the \textit{ceteris paribus} effect of growing $n$, we take random subsamples of the ionosphere data with varying size $n \in \{220, 260, 300\}$ and the full data set with $n = 350$. Figures~\ref{fig:iono-1} through \ref{fig:iono-4} show the respective results.  

It becomes apparent that with growing $n$ the differences between the performances of the two approximations diminishes. Already for small $n$

\clearpage

\begin{figure}[!h]
    \centering
    \includegraphics[scale=0.7]{Sample UAI 2023 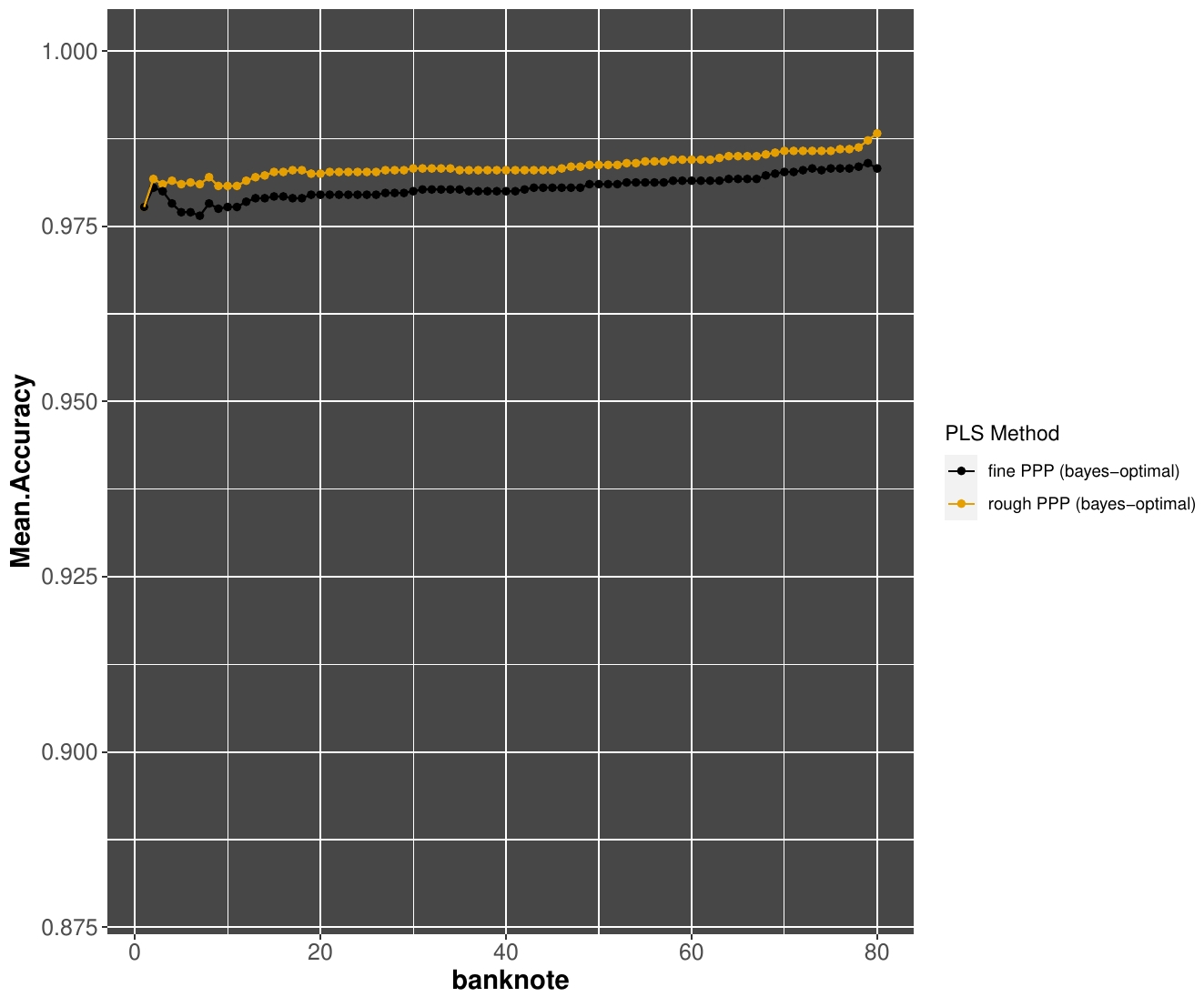}
    \caption{Approximations' performances on banknote data set ($n=200$, $q = 3$).}
    \label{fig:banknote}
\end{figure}

\clearpage
\newpage
\section{REPRODUCIBILITY AND OPEN SCIENCE}

The implementation of the proposed methods as well as
reproducible scripts for the experiments are provided in the following anonymous repository named \textbf{Bayesian-pls} (\say{\textit{Bayesian, please!}}): \url{https://anonymous.4open.science/r/Bayesian-pls}. Please follow the instructions on the Readme-file to reproduce the experiments. After the reviewing process, the whole (non-anonymous) repository will be made public.

\section{DATA SETS}
\label{sec:data-sets}

The following tables provide details on data sources as well as features and target variables of the eight real-world datasets from the UCI machine learning repository \cite{Dua:2019}.

\begin{table}[H] \centering \renewcommand*{\arraystretch}{1.1}\caption{Breast Cancer Data, Details: \cite{street1993nuclear}}
\begin{tabular}{p{0.290909090909091\textwidth}p{0.145454545454545\textwidth}p{0.363636363636364\textwidth}}
\hline
\hline
Name & Class & Values \\ 
\hline
target & factor & '0' '1' \\ 
radius\_mean & numeric & Num: 6.981 to 28.11 \\ 
texture\_mean & numeric & Num: 9.71 to 33.81 \\ 
perimeter\_mean & numeric & Num: 43.79 to 188.5 \\ 
area\_mean & numeric & Num: 143.5 to 2501 \\ 
smoothness\_mean & numeric & Num: 0.053 to 0.145 \\ 
compactness\_mean & numeric & Num: 0.019 to 0.311 \\ 
concavity\_mean & numeric & Num: 0 to 0.427 \\ 
concave\_points\_mean & numeric & Num: 0 to 0.201 \\ 
symmetry\_mean & numeric & Num: 0.117 to 0.304 \\ 
fractal\_dimension\_mean & numeric & Num: 0.05 to 0.097 \\ 
radius\_se & numeric & Num: 0.112 to 2.873 \\ 
texture\_se & numeric & Num: 0.36 to 4.885 \\ 
perimeter\_se & numeric & Num: 0.757 to 21.98 \\ 
area\_se & numeric & Num: 6.802 to 542.2 \\ 
smoothness\_se & numeric & Num: 0.002 to 0.031 \\ 
compactness\_se & numeric & Num: 0.002 to 0.106 \\ 
concavity\_se & numeric & Num: 0 to 0.396 \\ 
concave\_points\_se & numeric & Num: 0 to 0.053 \\ 
symmetry\_se & numeric & Num: 0.008 to 0.061 \\ 
fractal\_dimension\_se & numeric & Num: 0.001 to 0.03 \\ 
radius\_worst & numeric & Num: 7.93 to 36.04 \\ 
texture\_worst & numeric & Num: 12.02 to 49.54 \\ 
perimeter\_worst & numeric & Num: 50.41 to 251.2 \\ 
area\_worst & numeric & Num: 185.2 to 4254 \\ 
smoothness\_worst & numeric & Num: 0.071 to 0.223 \\ 
compactness\_worst & numeric & Num: 0.027 to 1.058 \\ 
concavity\_worst & numeric & Num: 0 to 1.252 \\ 
concave\_points\_worst & numeric & Num: 0 to 0.287 \\ 
symmetry\_worst & numeric & Num: 0.156 to 0.664 \\ 
fractal\_dimension\_worst & numeric & Num: 0.055 to 0.208\\ 
\hline
\hline
\end{tabular}
\end{table}

\begin{table}[H] \centering 
\tiny \renewcommand*{\arraystretch}{1.1}\caption{Sonar Data Set, Details: \cite{gorman1988analysis}}
\begin{tabular}{p{0.290909090909091\textwidth}p{0.145454545454545\textwidth}p{0.363636363636364\textwidth}}
\hline
\hline
Name & Class & Values \\ 
\hline
V1 & numeric & Num: 0.002 to 0.137 \\ 
V2 & numeric & Num: 0.001 to 0.234 \\ 
V3 & numeric & Num: 0.002 to 0.306 \\ 
V4 & numeric & Num: 0.006 to 0.426 \\ 
V5 & numeric & Num: 0.007 to 0.401 \\ 
V6 & numeric & Num: 0.01 to 0.382 \\ 
V7 & numeric & Num: 0.003 to 0.373 \\ 
V8 & numeric & Num: 0.005 to 0.459 \\ 
V9 & numeric & Num: 0.007 to 0.683 \\ 
V10 & numeric & Num: 0.011 to 0.711 \\ 
V11 & numeric & Num: 0.029 to 0.734 \\ 
V12 & numeric & Num: 0.024 to 0.706 \\ 
V13 & numeric & Num: 0.018 to 0.713 \\ 
V14 & numeric & Num: 0.027 to 0.997 \\ 
V15 & numeric & Num: 0.003 to 1 \\ 
V16 & numeric & Num: 0.016 to 0.999 \\ 
V17 & numeric & Num: 0.035 to 1 \\ 
V18 & numeric & Num: 0.038 to 1 \\ 
V19 & numeric & Num: 0.049 to 1 \\ 
V20 & numeric & Num: 0.066 to 1 \\ 
V21 & numeric & Num: 0.051 to 1 \\ 
V22 & numeric & Num: 0.022 to 1 \\ 
V23 & numeric & Num: 0.056 to 1 \\ 
V24 & numeric & Num: 0.024 to 1 \\ 
V25 & numeric & Num: 0.024 to 1 \\ 
V26 & numeric & Num: 0.092 to 1 \\ 
V27 & numeric & Num: 0.048 to 1 \\ 
V28 & numeric & Num: 0.028 to 1 \\ 
V29 & numeric & Num: 0.014 to 1 \\ 
V30 & numeric & Num: 0.061 to 1 \\ 
V31 & numeric & Num: 0.048 to 0.966 \\ 
V32 & numeric & Num: 0.04 to 0.931 \\ 
V33 & numeric & Num: 0.048 to 1 \\ 
V34 & numeric & Num: 0.021 to 0.965 \\ 
V35 & numeric & Num: 0.022 to 1 \\ 
V36 & numeric & Num: 0.008 to 1 \\ 
V37 & numeric & Num: 0.035 to 0.95 \\ 
V38 & numeric & Num: 0.038 to 1 \\ 
V39 & numeric & Num: 0.037 to 0.986 \\ 
V40 & numeric & Num: 0.012 to 0.93 \\ 
V41 & numeric & Num: 0.036 to 0.899 \\ 
V42 & numeric & Num: 0.006 to 0.825 \\ 
V43 & numeric & Num: 0 to 0.773 \\ 
V44 & numeric & Num: 0 to 0.776 \\ 
V45 & numeric & Num: 0 to 0.703 \\ 
V46 & numeric & Num: 0 to 0.729 \\ 
V47 & numeric & Num: 0 to 0.552 \\ 
V48 & numeric & Num: 0 to 0.334 \\ 
V49 & numeric & Num: 0 to 0.198 \\ 
V50 & numeric & Num: 0 to 0.082 \\ 
V51 & numeric & Num: 0 to 0.1 \\ 
V52 & numeric & Num: 0.001 to 0.071 \\ 
V53 & numeric & Num: 0 to 0.039 \\ 
V54 & numeric & Num: 0.001 to 0.035 \\ 
V55 & numeric & Num: 0.001 to 0.045 \\ 
V56 & numeric & Num: 0 to 0.039 \\ 
V57 & numeric & Num: 0 to 0.035 \\ 
V58 & numeric & Num: 0 to 0.044 \\ 
V59 & numeric & Num: 0 to 0.036 \\ 
V60 & numeric & Num: 0.001 to 0.044 \\ 
V61 & matrix & Num: 1 to 2\\ 
\hline
\hline
\end{tabular}
\end{table}

\begin{table}[!htbp] \centering \renewcommand*{\arraystretch}{1.1}\caption{Mushrooms Data Set, Details: \cite{schlimmer1987concept}}
\begin{tabular}{p{0.290909090909091\textwidth}p{0.145454545454545\textwidth}p{0.363636363636364\textwidth}}
\hline
\hline
Name & Class & Values \\ 
\hline
cap.diameter & numeric & Num: 0.71 to 54.6 \\ 
stem.height & numeric & Num: 0 to 28.33 \\ 
stem.width & numeric & Num: 0 to 52.22 \\ 
target & factor & '0' '1'\\ 
\hline
\hline
\end{tabular}
\end{table}

\begin{table}[!htbp] \centering \renewcommand*{\arraystretch}{1.1}\caption{Banknote Data Set, Details: \href{https://archive.ics.uci.edu/ml/datasets/banknote+authentication}{archive.ics.uci.edu/ml/datasets/banknote+authentication}}
\begin{tabular}{p{0.290909090909091\textwidth}p{0.145454545454545\textwidth}p{0.363636363636364\textwidth}}
\hline
\hline
Name & Class & Values \\ 
\hline
target & factor & '0' '1' \\ 
Length & numeric & Num: 213.8 to 216.3 \\ 
Left & numeric & Num: 129 to 131 \\ 
Right & numeric & Num: 129 to 131.1 \\ 
Bottom & numeric & Num: 7.2 to 12.7 \\ 
Top & numeric & Num: 7.7 to 12.3 \\ 
Diagonal & numeric & Num: 137.8 to 142.4\\ 
\hline
\hline
\end{tabular}
\end{table}

\begin{table}[!htbp] \centering \renewcommand*{\arraystretch}{1.1}\caption{Abalone Data Set, Details: \cite{waugh1995extending}}
\begin{tabular}{p{0.290909090909091\textwidth}p{0.145454545454545\textwidth}p{0.363636363636364\textwidth}}
\hline
\hline
Name & Class & Values \\ 
\hline
target & factor & '0' '1' \\ 
rings & numeric & Num: 4 to 29 \\ 
length & numeric & Num: 0.165 to 0.775 \\ 
weight & numeric & Num: 0.024 to 2.493 \\ 
height & numeric & Num: 0.04 to 0.24 \\ 
diameter & numeric & Num: 0.125 to 0.605 \\ 
shell\_weight & numeric & Num: 0.008 to 0.885\\ 
\hline
\hline
\end{tabular}
\end{table}

\begin{table}[!htbp] \centering \renewcommand*{\arraystretch}{1.1}\caption{Cars Data Set, Details: \cite{ezekiel1930methods}}
\begin{tabular}{p{0.290909090909091\textwidth}p{0.145454545454545\textwidth}p{0.363636363636364\textwidth}}
\hline
\hline
Name & Class & Values \\ 
\hline
wt & numeric & Num: 1.513 to 5.424 \\ 
qsec & numeric & Num: 14.5 to 22.9 \\ 
vs & factor & '0' '1' \\ 
\hline
\hline
\end{tabular}
\end{table}

\begin{table}[!htbp] \centering \renewcommand*{\arraystretch}{1.1}\caption{EEG Data Set, Details: \cite{zhang1995event}}
\begin{tabular}{p{0.290909090909091\textwidth}p{0.145454545454545\textwidth}p{0.363636363636364\textwidth}}
\hline
\hline
Name & Class & Values \\ 
\hline
V1 & numeric & Num: -2.035 to 1 \\ 
V2 & numeric & Num: -1.005 to 1 \\ 
V3 & numeric & Num: -0.912 to 1 \\ 
V4 & numeric & Num: -1.107 to 1 \\ 
V5 & numeric & Num: -1.078 to 1 \\ 
V6 & numeric & Num: -1.073 to 1 \\ 
V7 & numeric & Num: -1.651 to 1 \\ 
V8 & numeric & Num: -1.024 to 1 \\ 
V9 & numeric & Num: -1.864 to 1 \\ 
V10 & numeric & Num: -1.604 to 1 \\ 
V11 & numeric & Num: -0.883 to 1 \\ 
V12 & numeric & Num: -1.087 to 1 \\ 
target & factor & '0' '1'\\ 
\hline
\hline
\end{tabular}
\end{table}

\vfill

\begin{table}[H] \centering \renewcommand*{\arraystretch}{1.1}\caption{Ionosphere Data, Details: \cite{sigillito1989classification}}
\begin{tabular}{p{0.290909090909091\textwidth}p{0.145454545454545\textwidth}p{0.363636363636364\textwidth}}
\hline
\hline
Name & Class & Values \\ 
\hline
V1 & integer & Num: 0 to 1 \\ 
V3 & numeric & Num: -1 to 1 \\ 
V4 & numeric & Num: -1 to 1 \\ 
V5 & numeric & Num: -1 to 1 \\ 
V6 & numeric & Num: -1 to 1 \\ 
V7 & numeric & Num: -1 to 1 \\ 
V8 & numeric & Num: -1 to 1 \\ 
V9 & numeric & Num: -1 to 1 \\ 
V10 & numeric & Num: -1 to 1 \\ 
V11 & numeric & Num: -1 to 1 \\ 
V12 & numeric & Num: -1 to 1 \\ 
V13 & numeric & Num: -1 to 1 \\ 
V14 & numeric & Num: -1 to 1 \\ 
V15 & numeric & Num: -1 to 1 \\ 
V16 & numeric & Num: -1 to 1 \\ 
V17 & numeric & Num: -1 to 1 \\ 
V18 & numeric & Num: -1 to 1 \\ 
V19 & numeric & Num: -1 to 1 \\ 
V20 & numeric & Num: -1 to 1 \\ 
V21 & numeric & Num: -1 to 1 \\ 
V22 & numeric & Num: -1 to 1 \\ 
V23 & numeric & Num: -1 to 1 \\ 
V24 & numeric & Num: -1 to 1 \\ 
V25 & numeric & Num: -1 to 1 \\ 
V26 & numeric & Num: -1 to 1 \\ 
V27 & numeric & Num: -1 to 1 \\ 
V28 & numeric & Num: -1 to 1 \\ 
V29 & numeric & Num: -1 to 1 \\ 
V30 & numeric & Num: -1 to 1 \\ 
V31 & numeric & Num: -1 to 1 \\ 
V32 & numeric & Num: -1 to 1 \\ 
V33 & numeric & Num: -1 to 1 \\ 
V34 & numeric & Num: -1 to 1 \\ 
target & factor & '0' '1'\\ 
\hline
\hline
\end{tabular}
\end{table}

\vfill

\newpage

\vfill

\clearpage

\section{REFERENCES OF SUPPLEMENTARY MATERIAL}
\bibliographystyle{apalike}
\bibliography{literature}